\documentclass{article} % For LaTeX2e
\usepackage{iclr2027_conference,times}
\iclrfinalcopy
\usepackage{xcolor}

\usepackage{graphicx}
\usepackage{amsmath,amssymb}

\usepackage{amsmath,amsfonts,bm}

\def\eqref#1{equation~\ref{#1}}
\def\1{\bm{1}}

\DeclareMathAlphabet{\mathsfit}{\encodingdefault}{\sfdefault}{m}{sl}
\SetMathAlphabet{\mathsfit}{bold}{\encodingdefault}{\sfdefault}{bx}{n}

\usepackage{hyperref}
\usepackage{url}
\usepackage{enumitem}              % add this
\setlist[itemize]{leftmargin=*}

\title{A Model-Agnostic Physics-Guided Adapter for Few-Shot Transfer of Coastal Flood Prediction Models to Unseen Regions}

\author{Bilal Hassan, Areg Karapetyan and Samer Madanat \\
Division of Engineering\\
New York University Abu Dhabi\\
Abu Dhabi, UAE \\
\texttt{\{bilal.hassan, areg.karapetyan, samer.madanat\}@nyu.edu}
}

\begin{document}

\maketitle

\begin{abstract}

Deep learning (DL) surrogates can produce high-resolution coastal flood maps orders of magnitude faster than physics-based hydrodynamic simulators, yet transferring them to new coastal regions remains costly, since generating target-region data for fine-tuning typically requires numerous time-consuming simulations. To tackle this bottleneck, we introduce the Physics Adapter (PA), a compact, architecture-agnostic adaptation interface that enables efficient few-shot transfer of flood prediction models across diverse coastal regions. PA predicts peak water level through a differentiable wet/dry response that compares terrain elevation against a learned water level, and blends this physics-structured prediction with a data-driven branch through a learned gate. Unlike physics-informed formulations, PA imposes no PDE-residual or conservation losses and instead exploits elevation as an architectural inductive bias, adding a negligible number of trainable parameters. We integrate PA into 12 heterogeneous models, spanning graph, convolutional, Transformer, state-space, depth-foundation and diffusion models, and evaluate them on two coastal regions with markedly distinct geometries, topographies, and shoreline protection configurations. The performance of PA is benchmarked against a no-physics baseline, full fine-tuning, and standard parameter-efficient fine-tuning (PEFT) methods, considering both within-region generalization to unseen sea level rise (SLR) values and between-region transfer. 
In low-shot regime (K=3), and averaged over all backbones and transfer settings, adding PA reduces root mean square error (RMSE) by 11.5\% when only the output head is adapted on a frozen backbone, by 15.4\% when combined with PEFT methods, and by 22.9\% under full fine-tuning, compared to matched configurations without PA. Taken together, the findings of this work offer practitioners a concrete recipe for extending DL-based coastal flood predictors to new, data-scarce regions, thereby advancing scalable AI support for coastal adaptation planning.

\end{abstract}

\section{Introduction}
\label{sec:introduction}

Climate adaptation-aware coastal protection planning requires repeated prediction of how peak water level (PWL) responds to candidate shoreline protection configurations under different sea level rise (SLR) and forcing conditions. Physics-based high-fidelity simulators, such as Delft3D \citep{lesser2004delft3d}, can accurately simulate nearshore hydrodynamics, providing fine-grained estimates of depth, duration, and velocity of floods. However, due to prohibitively high computational cost, their direct adoption in large-scale coastal protection investigations, where each combination of protection configuration and SLR value requires a separate simulation, remains impractical~\citep{Jia2019InvestigationAnalysis}. Prior studies~\citep{hassan2026hess, karapetyan2026caspian, bian2025coastal} have demonstrated that learned surrogate models can dramatically reduce this computational burden by approximating the simulator's output, thereby replacing repeated hydrodynamic simulations with efficient inference.

%provide the reference simulations used for this purpose, but evaluating many protection and SLR values requires a separate numerical solve for each scenario \citep{karapetyan2026caspian, Jia2019InvestigationAnalysis}. Learned surrogate models reduce this computational burden by approximating the simulator's output, thereby replacing repeated hydrodynamic simulations with efficient inference. Previous work, such as the studies in~\citep{hassan2026hess, karapetyan2026caspian, bian2025coastal},  developed coastal PWL surrogates for San Francisco (SF) and Abu Dhabi (AD),

%Their value at deployment, however, depends on more than in-domain accuracy because a model trained for one coastline or forcing condition may be deployed in a target domain with different terrain, protection geometry, land-cover structure, or flood response.

%Previous Delft3D-based work developed coastal PWL surrogates for San Francisco (SF) and Abu Dhabi (AD), but showed that the performance can degrade sharply under domain shift \citep{hassan2026hess}. 

Existing surrogate models have been typically trained and evaluated for one coastline or forcing condition, and their accuracy can degrade sharply when applied to a different coastal region or forcing conditions outside the training range (Sec.~\ref{sec:results}; see also \citealp{Zhao2026}). This hinders their practical deployment, since generating sufficient training data for every new setting entails additional hydrodynamic simulations and substantial compute. Moreover, how well existing surrogates transfer across coastlines and SLR conditions, and how many target simulations are necessary to attain satisfactory performance, remains largely unexamined.

%This limits practical deployment, since retraining a surrogate for every new coastal setting requires additional hydrodynamic simulations and substantial compute. 

In this paper, we investigate whether pretrained coastal flood prediction models can be adapted to new regions and SLR conditions from only a few target simulations, and whether this can be achieved in an architecture-agnostic manner. The problem is challenging for two reasons. First, domain shift arises in different forms: geographic transfer changes terrain, protection geometry, land-cover structure, and flood response, whereas SLR transfer affects the forcing within the same region. An effective adaptation mechanism must handle both from only a handful of labeled examples. Second, achieving this in an architecture-agnostic manner is difficult, since the surrogate models can span various learned representations, from graph and mesh networks to vision, state-space and diffusion models. As \citet{lee2023surgical} illustrate, with limited target data the choice of which parameters to update matters, and the best choice depends on the type of shift. Parameter-efficient fine-tuning (PEFT) methods, such as LoRA \citep{hu2022lora}, BitFit \citep{benzaken2022bitfit}, and IA\(^3\) \citep{liu2022ia3}, restrict updates to selected parameters or inserted modules. These methods specify where and how a source model changes, but do not encode any flood-specific relation between PWL and the physical variables that remain observable in the target domain. In this context, we treat parameter efficiency and physical structure as distinct, potentially complementary components of adaptation. \looseness-2

%We therefore investigate whether target-domain performance can be recovered from only a few labeled target scenarios through a shared physics-guided adaptation strategy that applies across different backbone families. This is particularly challenging for two reasons. First, domain shift arises in different forms. Geographic transfer changes terrain, protection geometry, land-cover structure, and flood response, while SLR transfer changes the forcing within the same region. Second, the strategy must work across very different learned representations, including graph, vision, state-space, foundation-model, and diffusion architectures.

%We formalize this as supervised few-shot adaptation, in which only a small labeled support set of target scenarios is available. With limited target data, the choice of which parameters to update matters \citep{lee2023surgical}. Parameter-efficient fine-tuning (PEFT) methods, such as LoRA \citep{hu2022lora}, BitFit \citep{benzaken2022bitfit}, and IA\(^3\) \citep{liu2022ia3}, restrict updates to selected parameters or inserted modules. They specify where and how a source model changes, but do not encode any flood-specific relation between PWL and the physical variables that remain observable in the target domain. We therefore treat parameter efficiency and physical structure as distinct, potentially complementary components of adaptation.

A recent article by \citet{Daramolahen_2026} argues that transferable coastal flood models require inductive biases reflecting the underlying hydrodynamic processes, rather than solely relying on statistical similarity between source and target regions. In line with this view, we anchor our approach on terrain elevation, which is readily available for coastal regions and is directly linked to inundation extent and dynamics. More concretely, we introduce a lightweight module, termed the \emph{Physics Adapter} (PA), that combines elevation with the features of a pretrained surrogate model (hereafter, also referred to as backbone) to predict PWL. A physics-guided branch predicts PWL through a differentiable wet/dry response that compares terrain elevation against a learned water level, while a parallel data-driven branch captures effects that terrain alone does not explain. A learned gate combines the two predictions. As PA requires only backbone features and elevation, it can be attached to any architecture, with only the integration interface tailored to each backbone. Unlike physics-informed formulations such as PINNs \citep{raissi2019pinn}, PA does \emph{not} embed the shallow-water or Navier Stokes equations, minimize PDE residuals, or enforce mass or momentum conservation. Terrain elevation thus serves PA as an \emph{architectural} inductive bias rather than a \emph{governing-equation constraint}. \looseness-2

We instantiate PA across 12 diverse backbones and evaluate it on two coastal regions with markedly different geometries, topographies, and shoreline protection configurations, namely the coastal city of Abu Dhabi (AD) and the San Francisco (SF) Bay Area. Our main contributions are as follows:

\begin{itemize}[leftmargin=*]
    \item \textbf{A lightweight, architecture-agnostic physics-guided adapter:} PA injects terrain-based physical structure into pretrained flood surrogates without PDE-residual or conservation losses, while adding a negligible number of trainable parameters. \looseness-2
    \item \textbf{Extensive Evaluation:} We evaluate twelve backbones spanning graph, dense-vision, state-space, foundation, and diffusion models under bidirectional cross-region transfer (SF\(\leftrightarrow\)AD) and within-region SLR transfer, comparing ten adaptation regimes (full fine-tuning, head-only adaptation, and three PEFT methods, each with and without PA) under a matched protocol.
\item \textbf{Empirical evidence that physical structure complements parameter efficiency:} Averaged across backbones and transfer settings, every regime with PA outperforms every regime without it once a single target simulation is available, and adapting PA alone surpasses full fine-tuning without PA. At the level of individual backbones, a PA regime remains the best parameter-efficient choice for nine of the twelve models.
\end{itemize}

%At \(K{=}3\), PA reduces RMSE by up to 22.9\% relative to matched configurations, while preserving or improving in-domain accuracy for ten of the twelve backbones.

\section{Related Work}
\label{sec:related_work}

\paragraph{Learned surrogates for flood prediction.} For coastal domains, DL-based surrogates have been developed for predicting extreme storm surge under future climate scenarios \citep{longo2026stormsurge,rice2025surge, gharehtoragh2024evolving}, spatiotemporal flood dynamics \citep{bian2025coastal}, and tidal and riverine shallow-water dynamics \citep{riveracasillas2025mitonet}. The CASPIAN framework~\citep{karapetyan2026caspian} and its follow-up~\citep{hassan2026hess} predict PWL under shoreline protection for SF and AD across multiple SLR conditions, and their publicly released dataset serves as the data source for this work. Most of these surrogates, however, are trained and evaluated within a single region. Only a few studies have examined how coastal surrogates transfer to regions unseen during training. \citet{Zhao2026}, for instance, report zero-shot generalization of a storm-surge model to unseen bays along the \textit{same coastline}. Similar efforts for urban and riverine flooding adapt neural-operator surrogates to new catchments or forcing conditions via transfer learning \citep{xu2025urbanflood} or domain adaptation \citep{taghizadeh2026floodforecaster}. These studies, however, examine transfer within a single surrogate design. On the other hand, the present work investigates whether one physics-guided adaptation strategy, shared across multiple backbone model families, can recover target-domain performance from only a few labeled target scenarios, including across geographically and hydrodynamically distinct coastlines.

%Learned surrogates approximate numerical simulators by mapping input scenarios to spatial hydraulic fields. Operator learning offers one general formulation \citep{li2021fno}. Flood-specific examples include Fourier neural operators for inundation prediction \citep{sun2023floodfno}, hydraulics-informed and multi-scale graph models \citep{kazadi2024hydraulics,bentivoglio2025mswegnn}, and dense coastal models evaluated across extreme sea-level conditions \citep{bian2025coastal}. The CASPIAN studies developed vision-based PWL predictors under shoreline protection for Abu Dhabi and San Francisco across multiple SLR conditions, and publicly released the Delft3D datasets used in this work \citep{hassan2026hess}. Some recent work also addresses transfer. Deep neural operators have been fine-tuned or domain-adapted across urban-flood scenarios \citep{xu2025urbanflood}, and FloodForecaster combines a geometry-informed neural operator with domain adaptation to transfer to new regions \citep{taghizadeh2026floodforecaster}. These studies show that transfer is possible within a single surrogate design. We examine whether a physics-guided adaptation strategy, shared across multiple backbone model families, can robustly recover target-domain performance when only a limited number of labeled target scenarios are available.

\vspace*{-5pt}
\paragraph{Physics-guided learning.}
Physical constraints can enter a learned surrogate through different mechanisms. PINNs impose governing equations through residual-based objectives \citep{raissi2019pinn}, and physics-informed neural operators combine operator learning with PDE constraints \citep{li2024pino}. Flood-specific models can encode more domain-specific hydraulic structure. HydroGraphNet \citep{taghizadeh2025hydrographnet} includes mass conservation in its training objective, whereas hydraulics-informed message passing derives graph interactions from shallow-water structure \citep{kazadi2024hydraulics}. Beyond flooding, physical structure can also be built into the architecture itself, as in ClimODE \citep{verma2024climode}, which encodes advection within continuous-time weather dynamics. Closest to our setting, GeoAda-PINN \citep{zhu2026geoadapinn} freezes a PINN backbone and updates compact geometry-conditioned adapters to handle geometric changes. In contrast, the proposed adapter uses a lighter architectural inductive bias based on terrain elevation and a differentiable wet/dry response, without any governing-equation loss or conservation guarantee. Moreover, PA employs a consistent formulation across heterogeneous backbone families, whereas GeoAda-PINN is tied to a single PINN architecture.

%Physical constraints can enter a learned surrogate through different mechanisms. PINNs impose governing equations through residual-based objectives \citep{raissi2019pinn}, and physics-informed neural operators combine operator learning with PDE constraints \citep{li2024pino}. Flood-specific models can encode stronger hydraulic structure. HydroGraphNet \citep{taghizadeh2025hydrographnet} includes mass conservation in its training objective, whereas hydraulics-informed message passing derives graph interactions from shallow-water structure \citep{kazadi2024hydraulics}. Other models build physical structure into the architecture itself. For instance, ClimODE \citep{verma2024climode} encodes advection within continuous-time weather dynamics. Closest to our setting, GeoAda-PINN \citep{zhu2026geoadapinn} freezes a PINN backbone and updates compact geometry-conditioned adapters to handle geometric changes. In contrast, the proposed PA uses a lighter architectural inductive bias based on terrain elevation and a differentiable wet/dry response, without any governing-equation loss or conservation guarantee, and it employs a consistent formulation across heterogeneous backbone families, rather than being confined to a single model.
\vspace*{-5pt}
\paragraph{Adaptation under distribution shift and PEFT.}
When labeled target data are scarce, a key question is which source parameters should be updated. Surgical fine-tuning shows that the effective subset can depend on the type of distribution shift \citep{lee2023surgical}. Unsupervised test-time methods instead adapt from unlabeled target batches during inference \citep{wang2021tent}. This differs from the proposed supervised few-shot setting, where labels are available for the target support scenarios. PEFT controls the optimization subspace through mechanisms such as bottleneck adapters \citep{houlsby2019adapters}, low-rank weight updates \citep{hu2022lora}, bias-only tuning \citep{benzaken2022bitfit}, and activation scaling \citep{liu2022ia3}. Architecture-aware PEFT has also been studied for state-space models \citep{yoshimura2025mambapeft}, and F-Adapter extends this line to large neural-operator models for scientific machine learning \citep{zhang2025fadapter}. These approaches alter the parameterization of adaptation, whereas the PA instead adds elevation-conditioned structure to the prediction interface. We therefore evaluate PA and PEFT both separately and in combination across heterogeneous neural surrogates. \looseness-2

\section{Method}
\label{sec:method}

This section details the proposed approach and its use for target adaptation. We separate how each backbone represents a flood scenario from how that representation is converted into elevation-conditioned PWL, so that this conversion can be shared across all twelve backbones. Sec.~\ref{sec:problem} defines the prediction problem and this separation, Sec.~\ref{sec:physics_adapter} describes the adapter, Sec.~\ref{sec:instantiation} explains how it is attached to each backbone, and Sec.~\ref{sec:adaptation} defines source training and the adaptation regimes.

\subsection{Problem Formulation}
\label{sec:problem}

A prediction domain is a pair \(d=(r,\lambda)\) of a geographic region \(r\) and a forcing condition \(\lambda\), which we vary through SLR. For region \(r\), let \(\Omega_r\) denote the discrete prediction sites and \(\Omega_r^{v}\subseteq\Omega_r\) the valid sites. A flood scenario has input \(\mathbf{x}\) and target \(\mathbf{y}=\{y_i\}_{i\in\Omega_r^{v}}\), where \(y_i\geq 0\) is the simulated PWL. Raster backbones take a four-channel \(1024\times1024\) tensor \(\mathbf{x}_{\mathrm{grid}}=[\mathbf{c}^{(\mathbf{s})},\mathbf{z},\boldsymbol{\ell},\mathbf{v}]\) containing scenario-dependent protection status, the DEM, land cover, and a binary validity mask \(v_i\in\{0,1\}\). Graph and mesh backbones encode the same variables as node features on the valid sites, so the mask is implicit (Appendix~\ref{app:data_construction}). In every case, the PA receives the raw DEM value \(z_i\) in physical units, rather than a normalized or embedded version recovered from backbone features.

\begin{figure*}[t]
    \centering
    \includegraphics[width=\linewidth]{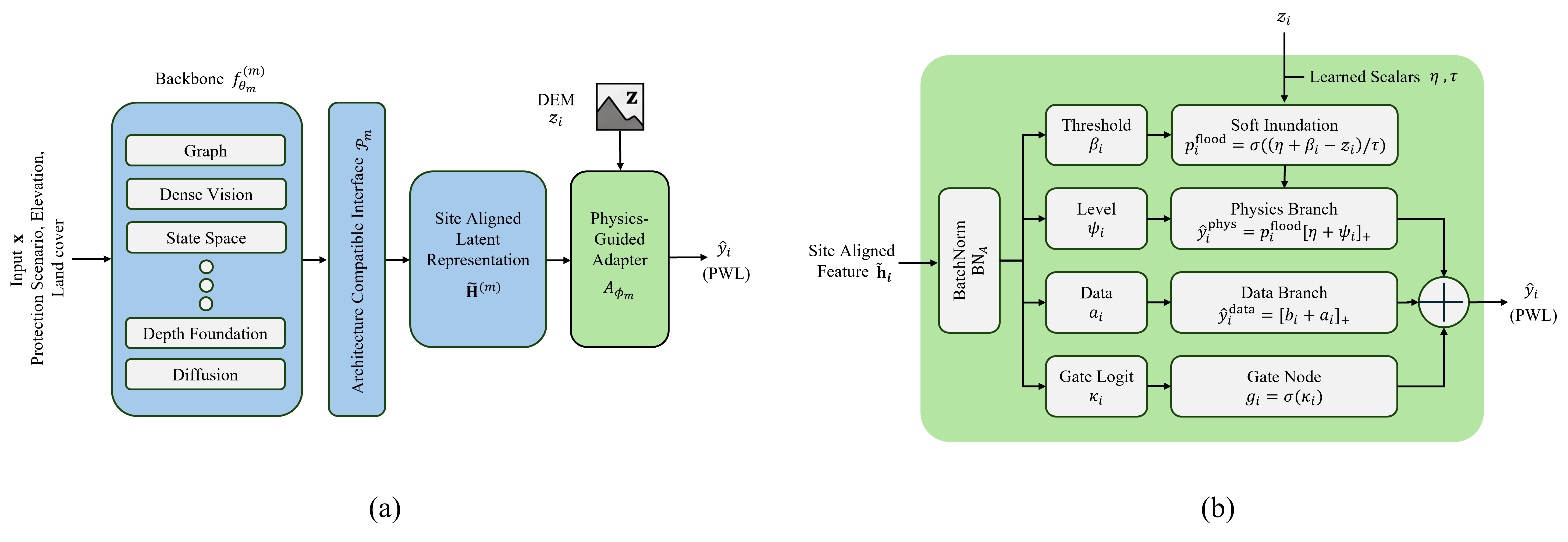}
    \caption{Backbone-agnostic Physics Adapter. (a)~Each backbone exposes features aligned with the PWL prediction sites through its own interface, and the shared adapter produces the prediction. (b)~Inside the adapter, raw terrain elevation enters the physics-guided branch through a soft wet/dry response, a parallel data-driven branch captures effects that terrain alone does not explain, and a learned site-wise gate combines the two. Feature extraction is architecture-specific, while terrain conditioning, the two branches, gated fusion, and target adaptation are shared.}
    \label{fig:method_overview}
\end{figure*}

Each of the twelve backbones \(m\in\mathcal{M}\) (Sec.~\ref{sec:benchmarks}) defines a native representation, a site-alignment interface \(\mathcal{P}_m:\mathcal{H}_m\rightarrow\mathbb{R}^{|\Omega_r|\times C_m}\), and the shared adapter computation
\begin{equation}
\begin{aligned}
\mathbf{h}^{(m)} &= f_{\theta_m}^{(m)}(\mathbf{x}),\\
\widetilde{\mathbf{H}}^{(m)} &= \mathcal{P}_m\bigl(\mathbf{h}^{(m)}\bigr),
\qquad \widetilde{\mathbf{h}}^{(m)}_i\in\mathbb{R}^{C_m},\\
\widehat{\mathbf{y}}^{(m)} &= A_{\phi_m}\bigl(\widetilde{\mathbf{H}}^{(m)},\mathbf{z},\mathbf{v}\bigr).
\end{aligned}
\label{eq:pipeline}
\end{equation}
The interface \(\mathcal{P}_m\) (attachment point, tensor layout, feature dimension, decoder, interpolation, and projection) differs across backbones. Its parameters are grouped with the backbone parameters \(\theta_m\), except in the diffusion case, where the post-decoder stem belongs to the adapter (Sec.~\ref{sec:instantiation}). The adapter parameters are \(\phi_m\), so the full PA model has parameters \(\Theta^{(m)}=\theta_m\cup\phi_m\). Here \(\theta_m\) determines how a scenario is represented, and \(\phi_m\) determines how that representation is converted into elevation-conditioned PWL. Figure~\ref{fig:method_overview} summarizes the formulation.

\subsection{Physics-Guided Adapter}
\label{sec:physics_adapter}

\paragraph{Adapter heads.}
The site-aligned features are first normalized by a BatchNorm layer, and four lightweight representation-compatible functions then produce a threshold correction ($\beta_i$), a water-level correction ($\psi_i$), a data-branch correction ($a_i$), and a gate logit ($\kappa_i$),
\begin{equation}
\begin{aligned}
\mathbf{h}_i &= \operatorname{BN}^{(m)}_{A}\bigl(\widetilde{\mathbf{h}}^{(m)}_i\bigr),\\
\beta_i = f_{\beta}^{(m)}(\mathbf{h}_i),\quad
\psi_i &= f_{\psi}^{(m)}(\mathbf{h}_i),\quad
a_i = f_{a}^{(m)}(\mathbf{h}_i),\quad
\kappa_i = f_{g}^{(m)}(\mathbf{h}_i).
\end{aligned}
\label{eq:adapter_fields}
\end{equation}
For dense feature maps, the heads are pointwise \(1\times1\) convolutions, and for node-aligned backbones they are node-wise two-layer multilayer perceptrons (MLPs). The running statistics of \(\operatorname{BN}^{(m)}_{A}\) are part of the adapter state and are recalibrated during target adaptation (Sec.~\ref{sec:adaptation}). Backbone-specific settings, including affine BatchNorm parameters and head dropout are listed in Appendix~\ref{app:architecture_interfaces}.

\paragraph{Elevation-conditioned inundation.}
A learned scalar \(\eta\in\mathbb{R}\) defines a reference water level, and a second scalar \(\vartheta\) parameterizes a positive transition temperature. The local inundation response is
\begin{equation}
p_i^{\mathrm{flood}}
=
\sigma\!\left(\frac{\eta+\beta_i-z_i}{\tau}\right),
\qquad
\tau=\max\bigl\{\exp(\vartheta),\,10^{-3}\bigr\},
\label{eq:flood_probability}
\end{equation}
where \(z_i\) is the raw DEM elevation and \(\sigma\) is the logistic sigmoid. The correction \(\beta_i\) shifts the effective inundation threshold based on the backbone features, and \(\tau\) controls how sharp the wet/dry transition is. Both \(\eta\) and \(\vartheta\) are learned, with region-specific initialization given in Appendix~\ref{app:architecture_interfaces}.

\paragraph{Physics and data branches.}
The two prediction branches are
\begin{equation}
\widehat{y}^{\mathrm{phys}}_i
=
p_i^{\mathrm{flood}}\,\bigl[\eta+\psi_i\bigr]_+,
\qquad
\widehat{y}^{\mathrm{data}}_i
=
\bigl[b_i^{(m)}+a_i\bigr]_+,
\label{eq:branches}
\end{equation}
where \([\,\cdot\,]_+=\max(\cdot,0)\). The correction \(\psi_i\) adjusts the water level separately from the threshold shift \(\beta_i\) in Eq.~(\ref{eq:flood_probability}), so the physics-guided branch ties PWL to absolute terrain elevation even when the backbone does not preserve DEM units. For eleven backbones \(b_i^{(m)}=0\), and the data branch is a direct non-negative prediction. For the diffusion backbone, \(b_i^{(m)}\) is the generated base PWL, so the data branch predicts a residual around the generated map (Sec.~\ref{sec:instantiation}).

\paragraph{Gated fusion and masking.}
A site-wise gate mixes the branches, and invalid raster cells are removed from the output,
\begin{equation}
g_i=\sigma(\kappa_i),
\qquad
\widetilde{y}_i
=
g_i\,\widehat{y}^{\mathrm{phys}}_i+(1-g_i)\,\widehat{y}^{\mathrm{data}}_i,
\qquad
\widehat{y}_i=v_i\,\widetilde{y}_i.
\label{eq:fusion}
\end{equation}
For graph and mesh backbones, nodes already coincide with prediction sites, so masking is implicit. The gate is initialized toward the physics-guided branch with a logit of \(+3\), so that \(g_i=\sigma(3)\approx0.95\) at the start of training while the gate gradient remains large enough for the model to shift weight toward the data branch where needed. This value is a fixed design choice rather than a tuned hyperparameter (Appendix~\ref{app:architecture_interfaces}).

Eqs.~(\ref{eq:flood_probability})--(\ref{eq:fusion}) thus act as an elevation-conditioned \emph{architectural inductive bias}, \emph{not} a hydrodynamic solver, and add no PDE-residual or conservation constraint. In the same sense, the PA is not a new backbone, since \(\phi_m\) only converts an exposed representation into PWL. It is also distinct from parameter-efficient fine-tuning (PEFT) methods such as LoRA, BitFit and IA\(^3\), which modify selected weights or activations inside an existing model. The two can therefore be used separately or together, and the adaptation regimes in Sec.~\ref{sec:adaptation} compare both options.

\subsection{Backbone-Compatible Instantiation}
\label{sec:instantiation}

The interfaces \(\mathcal{P}_m\) align features with the prediction sites but do not make the architectures identical, so tensor shape, insertion depth, feature dimension, and PEFT target modules differ across backbones. Node-aligned backbones apply the adapter node-wise, dense backbones use pointwise heads on grid-aligned decoder features after a projection where needed, and the diffusion backbone passes its decoded base PWL map through an adapter-owned stem. In all cases the raw DEM bypasses the backbone and enters Eq.~(\ref{eq:flood_probability}) directly, so the PA is model-agnostic in its formulation rather than in its placement inside each network (Appendix~\ref{app:architecture_interfaces}, Table~\ref{tab:backbone_interfaces}).

\subsection{Source Training and Target Adaptation}
\label{sec:adaptation}

\paragraph{Objective.}
Deterministic backbones use masked mean squared error over valid sites. PA models add a penalty that discourages the gate from collapsing onto the data-driven branch,
\begin{equation}
\begin{aligned}
\mathcal{L}_{\mathrm{pred}}
&=
\frac{\sum_{n\in\mathcal{B}}\sum_{i\in\Omega_r} v_{n,i}\bigl(\widehat{y}_{n,i}-y_{n,i}\bigr)^2}
     {\max\bigl(1,\;\sum_{n\in\mathcal{B}}\sum_{i\in\Omega_r} v_{n,i}\bigr)},\\
\mathcal{L}
&=
\mathcal{L}_{\mathrm{pred}}
+
\mathbb{I}_{\mathrm{PA}}\,
\frac{\lambda_g}{|\Omega_A|}\sum_{i\in\Omega_A}(1-g_i),
\qquad
\lambda_g=0.1,
\end{aligned}
\label{eq:objective}
\end{equation}
where \(\mathcal{B}\) is a mini-batch, \(\mathbb{I}_{\mathrm{PA}}=1\) only for PA models, and \(\Omega_A\) is the set of sites on which the gate is defined (Appendix~\ref{app:architecture_interfaces}). For node-based models, \(\mathcal{L}_{\mathrm{pred}}\) reduces to standard MSE over valid nodes. The diffusion backbone keeps its native diffusion objective alongside the map-level objective, with the decoded base PWL map detached from the diffusion computation (Appendix~\ref{app:architecture_interfaces}).

\paragraph{Source training.}
For each backbone, the PA source model is trained jointly on the source domain,
\begin{equation}
(\theta_{m,s}^{\star},\phi_{m,s}^{\star})
=
\arg\min_{\theta_m,\phi_m}\;
\mathbb{E}_{(\mathbf{x},\mathbf{y})\sim\mathcal{D}^{s}_{\mathrm{train}}}
\bigl[\mathcal{L}(\mathbf{x},\mathbf{y};\theta_m,\phi_m)\bigr],
\label{eq:source_pa}
\end{equation}
so PA-only adaptation starts from an adapter trained together with its source representation rather than a newly initialized module. We also train a raw source model, in which the PA is replaced by a direct prediction head with parameters \(\chi_m\), trained on \(\mathcal{L}_{\mathrm{pred}}\) alone, which serves as the no-PA control (Appendix~\ref{app:architecture_interfaces}). We write \(\Theta^{(m)\star}_{s}\) for the selected source parameters of either model.

\paragraph{Target adaptation.}
Let \(\mathcal{S}_{K}^{t}=\{(\mathbf{x}^{t}_j,\mathbf{y}^{t}_j)\}_{j=1}^{K}\) be the labeled target support set, disjoint from the held-out target test set. An adaptation regime \(a\) specifies a trainable subset \(\Theta_a^{(m)}\) and is optimized from the corresponding source checkpoint,
\begin{equation}
\vspace*{-5pt}
\widehat{\Theta}_{a,K}^{(m)}
=
\arg\min_{\Theta_a^{(m)}}\;
\mathcal{L}_{t}\bigl(\mathcal{S}_{K}^{t};\,\Theta_a^{(m)},\,\Theta_{\neg a}^{(m),\mathrm{frozen}}\bigr),
\label{eq:target_optimization}
\end{equation}
where \(\mathcal{L}_{t}\) is the objective of Eq.~(\ref{eq:objective}) evaluated on \(\mathcal{S}_{K}^{t}\). For PEFT regimes, \(\Theta_a^{(m)}\) also includes the injected PEFT parameters \(\xi_{m,a}\). The ten regimes and their trainable sets are defined in Sec.~\ref{sec:benchmarks}. Frozen components are frozen in both parameters and internal state, and for every PA regime with \(K>0\), adaptation starts with a gradient-free recalibration of the adapter BatchNorm on the support inputs (Appendix~\ref{app:adaptation_details}). At \(K=0\), no recalibration or gradient update is performed, so \(\widehat{\Theta}^{(m)}_{a,0}=\Theta^{(m)\star}_{s}\) and zero-shot results directly evaluate the source checkpoint.

We consider two shifts. Geographic transfer changes the region (\(r_s\neq r_t\)), whereas SLR transfer keeps the region fixed and changes only the forcing (\(r_s=r_t\), \(\lambda_s\neq\lambda_t\)). Transfer directions, support sizes, and the adaptation budget are given in Sec.~\ref{sec:protocols}.

\section{Experimental Setup}
\label{sec:setup}

\subsection{Datasets and Representations}
\label{sec:datasets}
We use the publicly released coastal-flood simulations of the CASPIAN studies for AD and SF \citep{karapetyan2026caspian,hassan2026hess}, produced with Delft3D under defined SLR, tidal forcing, and binary shoreline-protection configurations (Appendix~\ref{app:data_construction}). Each protection scenario protects a subset of \(N_r\) operational landscape units (OLUs), with \(N_{\mathrm{SF}}=30\) and \(N_{\mathrm{AD}}=17\). The regional datasets contain 285 SF scenarios at \(1.0\,\mathrm{m}\) SLR and 142 AD scenarios at \(0.5\,\mathrm{m}\) SLR, and two SF SLR-transfer targets at \(0.5\,\mathrm{m}\) and \(1.5\,\mathrm{m}\) contain 32 scenarios each. Each retained coastal location carries a scenario-dependent protection status derived from simulated single-OLU responses, together with elevation from Copernicus DEM GLO-30 \citep{copernicusdem2022} and land cover from ESA WorldCover \citep{zanaga2022worldcover}, which we sample at every location since the original datasets do not include them. Dense backbones use the \(1024\times1024\) tensor of Sec.~\ref{sec:problem}, and graph backbones use a graph over the same locations whose edges follow the hydrodynamic mesh. Both are encodings of the same samples rather than separate datasets (Appendix~\ref{app:data_construction}, Figure ~\ref{fig:data_pipeline}).

\subsection{Benchmark Models and Comparison Methods}
\label{sec:benchmarks}

The twelve backbones cover clearly different model families. These are graph and mesh models (GCN \citep{kipf2017gcn}, GAT \citep{velickovic2018gat}, and MeshGraphNet (MGN) \citep{pfaff2021meshgraphnets}), scientific attention over physical sites (Transolver++ \citep{luo2025transolverpp}), dense vision models (CASPIAN \citep{karapetyan2026caspian}, ConvNeXt~V2 \citep{woo2023convnextv2}, MaxViT \citep{tu2022maxvit}, and Swin Transformer~V2 \citep{liu2022swinv2}), a visual state-space model (VM-UNet with a VMamba encoder--decoder \citep{ruan2024vmunet,liu2024vmamba}), pretrained depth-foundation models (Depth Anything~V2 \citep{yang2024depthanythingv2} and Depth Pro \citep{bochkovskiy2025depthpro}), and a conditional diffusion model (ControlNet \citep{zhang2023controlnet}). The depth-foundation models are adapted to PWL through task-specific input and feature interfaces, not by reinterpreting their depth outputs. Table~\ref{tab:backbone_interfaces} in Appendix~\ref{app:architecture_interfaces} lists the representation each backbone passes to the PA.

Each backbone has two matched source models. The \emph{PA source} model trains the backbone and PA jointly (Eq.~(\ref{eq:source_pa})), and the \emph{raw source} model replaces the PA with a direct prediction head \(\chi_m\). We compare ten target-adaptation regimes per backbone. Full fine-tuning is run with the PA (FT+PA) and without it (FT). Partial fine-tuning without PA (NPA) trains only \(\chi_m\), and PA-only adaptation (PA) trains only \(\phi_m\). Three PEFT methods, namely LoRA \citep{hu2022lora} with rank \(r=8\), BitFit \citep{benzaken2022bitfit}, and IA\(^3\) \citep{liu2022ia3}, train only their injected parameters \(\xi_{m,a}\), both on their own (PEFT) and combined with the PA (PEFT+PA). Table~\ref{tab:adaptation_modes_full} in Appendix~\ref{app:adaptation_details} gives the trainable and frozen parameters of each regime, along with the PEFT formulations and insertion sites. All regimes share the same scenario manifests, support sets, test sets, seeds, and metric code, while batch sizes, PEFT insertion sites, and parameter counts remain architecture-specific.

\vspace*{-25pt}
\subsection{Splits and Transfer Protocols}
\label{sec:protocols}

All backbones and seeds share one fixed scenario split, stratified by protection level into approximately \(60/20/20\) train, validation, and test sets, and model selection uses the validation split only. Geographic transfer is evaluated in both directions between SF and AD with \(K\in\{0,1,3,5,10\}\), and SLR transfer from \(\mathrm{SF}_{1.0}\) to \(\mathrm{SF}_{0.5}\) and \(\mathrm{SF}_{1.5}\) with \(K\in\{0,1,3,5,10\}\). For each \(K>0\), eight support draws are adapted independently from the source checkpoint with a fixed budget of 50 support passes and no target validation, and all are evaluated on the same held-out target test set (Appendix~\ref{app:splits}, Table~\ref{tab:protocol}).

\vspace*{-5pt}
\subsection{Hyperparameter Optimization , Training And Evaluation Metrics}
\label{sec:hpo}

For the details on hyperparameter optimization and training, we refer the reader to Appendix~\ref{app:training_details}.

We report the mean absolute error (MAE), RMSE, the coefficient of determination (\(R^2\)), the dry-point accuracy (\(\mathrm{Acc}_0\)), the relative total absolute error (RTAE), and the error exceedance rates \(\delta_{0.5}\) and \(\delta_{0.1}\). All metrics are computed at the same retained locations for every model, and their definitions and aggregation are given in Appendix~\ref{app:evaluation_details}.

%\subsection{Evaluation Metrics}
%\label{sec:metrics}

\section{Results}
\label{sec:results}

\subsection{In-Domain Prediction}
\label{sec:results_indomain}

Table~\ref{tab:main_summary} reports in-domain RMSE and \(R^2\) for each backbone, averaged over the SF and AD test sets. For every backbone we trained both a PA source model and a raw source model, and the table shows whichever of the two had the lower combined RMSE. The PA source model is the better of the two for ten of the twelve backbones. The two exceptions are GAT and Transolver++, where the raw model is slightly better in domain. The PA therefore does not cost in-domain accuracy in most cases, even though its main purpose is transfer.

VM-UNet is the most accurate backbone in both regions, with a combined RMSE of 0.054\,m and \(R^2=0.966\). Swin~V2, MaxViT, and CASPIAN follow closely, and the two depth-foundation models come next. The graph and operator backbones reach a similar \(R^2\) of about 0.92, but their absolute errors are several times larger than those of the dense backbones, so we compare RMSE mainly within each family. Across all backbones, errors are lower in SF than in AD, which is consistent with the stronger wave forcing and run-up in the AD simulations (Appendix~\ref{app:data_hydrodynamics}). Full results for all seven metrics, per region, are given in Appendix~\ref{app:results_indomain}.

\begin{table*}[t]
\centering
\footnotesize
\setlength{\tabcolsep}{2pt}
\caption{Per-backbone results. In-domain values report the better PA or raw source model over SF and AD. Transfer results report the lowest mean RMSE over all \(K\) and all four settings, with and without full fine-tuning. Values are mean \(\pm\) std over three seeds. Bold shows the best results.}
\label{tab:main_summary}

\resizebox{\textwidth}{!}{%
\begin{tabular}{l c c c| c c c c}
\hline
& \multicolumn{3}{c}{\textbf{In-domain}} 
& \multicolumn{4}{c|}{\textbf{Transfer (mean over \(K\))}} \\
\textbf{Backbone} 
& \textbf{Source} 
& \textbf{RMSE (m)} 
& \(\boldsymbol{R^2}\) 
& \textbf{Best regime} 
& \textbf{RMSE} 
& \textbf{Best without full FT} 
& \textbf{RMSE} \\
\hline
VM-UNet & PA & \textbf{0.0542} \(\pm\) 0.0015 & \textbf{0.9660} \(\pm\) 0.0008 & FT+PA & 0.1421 & LoRA & \textbf{0.1467} \\
Swin~V2 & PA & 0.0630 \(\pm\) 0.0025 & 0.9582 \(\pm\) 0.0055 & FT+PA & 0.1635 & LoRA+PA & 0.1662 \\
MaxViT & PA & 0.0639 \(\pm\) 0.0033 & 0.9564 \(\pm\) 0.0039 & FT+PA & 0.1663 & LoRA+PA & 0.1749 \\
CASPIAN & PA & 0.0686 \(\pm\) 0.0036 & 0.9560 \(\pm\) 0.0060 & FT+PA & \textbf{0.1388} & LoRA+PA & 0.1477 \\
Depth Pro & PA & 0.0713 \(\pm\) 0.0046 & 0.9522 \(\pm\) 0.0005 & FT+PA & 0.2245 & LoRA & 0.2423 \\
Depth Anything~V2 & PA & 0.0747 \(\pm\) 0.0017 & 0.9503 \(\pm\) 0.0052 & FT+PA & 0.2200 & LoRA & 0.2340 \\
ControlNet & PA & 0.0754 \(\pm\) 0.0030 & 0.9244 \(\pm\) 0.0096 & FT+PA & 0.1637 & LoRA+PA & 0.1722 \\
ConvNeXt~V2 & PA & 0.0893 \(\pm\) 0.0028 & 0.9217 \(\pm\) 0.0157 & FT+PA & 0.1648 & LoRA+PA & 0.1663 \\
\hline
MGN & PA & 0.2990 \(\pm\) 0.0034 & 0.9266 \(\pm\) 0.0011 & FT+PA & 0.6199 & IA\(^3\)+PA & 0.6449 \\
GAT & Raw & 0.3106 \(\pm\) 0.0019 & 0.9235 \(\pm\) 0.0007 & FT+PA & 0.4950 & BitFit+PA & 0.5760 \\
Transolver++ & Raw & 0.3417 \(\pm\) 0.0052 & 0.9215 \(\pm\) 0.0079 & LoRA+PA & 0.7125 & LoRA+PA & 0.7125 \\
GCN & PA & 0.3418 \(\pm\) 0.0020 & 0.9121 \(\pm\) 0.0009 & FT+PA & 0.5209 & BitFit+PA & 0.6456 \\
\hline
\end{tabular}%
}
\end{table*}

\subsection{Few-Shot Transfer}
\label{sec:results_transfer}

\begin{figure*}[t]
    \centering
    \includegraphics[width=\textwidth]{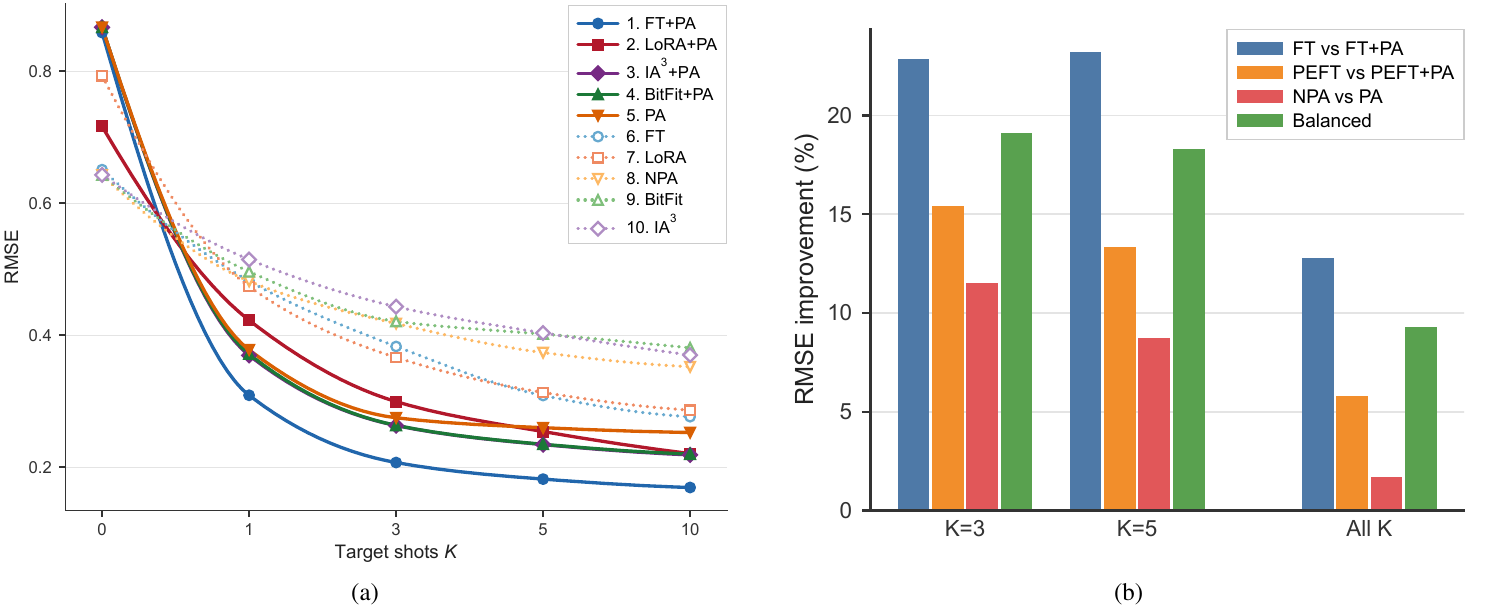}
\caption{Few-shot transfer averaged over 12 backbones and four settings. (a) RMSE across \(K\) for the ten adaptation regimes, ranked by mean RMSE. Solid lines use PA; dotted lines do not. (b) RMSE reduction from PA at \(K=3\), \(K=5\), and across all \(K\). ``Balanced'' averages the three pairings.}
\label{fig:transfer_main}
\end{figure*}

We first compare the ten regimes averaged over all twelve backbones and all four transfer settings (SF\(\rightarrow\)AD, AD\(\rightarrow\)SF, SF\(_{1.0}\)\(\rightarrow\)SF\(_{0.5}\), and SF\(_{1.0}\)\(\rightarrow\)SF\(_{1.5}\)). Since the SLR targets stop at \(K=10\), this comparison uses \(K\leq10\). Figure~\ref{fig:transfer_main}a shows the resulting RMSE curves, with regimes ranked by their mean RMSE over \(K\).

\paragraph{Regimes with the PA.}
The five regimes that include the PA take the top five places, and the five regimes without it take the bottom five. From \(K=1\) onward, every PA regime has a lower RMSE than every non-PA regime at every value of \(K\). FT+PA is the best regime overall, reducing RMSE from 0.856\,m at \(K=0\) to 0.169\,m at \(K=10\). The more useful comparison for practice is PA-only adaptation, which updates only the adapter parameters \(\phi_m\). It has a lower RMSE than full fine-tuning without the PA at every \(K\geq1\), even though FT updates the whole backbone. Among the PEFT methods, adding the PA lowers the error for LoRA, BitFit, and IA\(^3\) alike, and the three PEFT+PA regimes end close to each other at \(K=10\).

\paragraph{Zero-shot behavior.}
At \(K=0\) the ordering is reversed, and the raw source models transfer better than the PA source models (RMSE of about 0.674\,m against 0.856\,m). A likely reason is that the reference water level \(\eta\) in Eq.~(\ref{eq:flood_probability}) is learned for the source region, so without any target data the terrain comparison is made against the wrong water level. A single labeled target scenario is enough to reverse this, and at \(K=1\) all PA regimes are already ahead. In practice, the PA should be used with at least one target simulation, and zero-shot use requires care.

\paragraph{Where the gain comes from.}
Figure~\ref{fig:transfer_main}b isolates the effect of the PA by comparing each regime with its matched counterpart, namely FT with FT+PA, each PEFT method with its PEFT+PA version, and NPA with PA. The PA lowers RMSE by 22.87\% for full fine-tuning, 15.39\% for PEFT, and 11.51\% for head-only adaptation at \(K=3\), and by 12.77\%, 5.78\%, and 1.68\% averaged over all \(K\). The NPA against PA pairing is the cleanest test, since both regimes train only a small head on a frozen backbone and differ only in whether that head is conditioned on terrain. This gain points to the terrain conditioning as the main source of the improvement, although the two heads also differ somewhat in size (Appendix~\ref{app:evaluation_details}).

\paragraph{Per-backbone results.}
The last four columns of Table~\ref{tab:main_summary} show the best regime for each backbone. FT+PA is the best choice for eleven of the twelve backbones, and LoRA+PA is best for Transolver++. Because full fine-tuning is the most expensive option, we also report the best regime when both full fine-tuning regimes are excluded. In this case, a regime with the PA is still best for nine of the twelve backbones. LoRA+PA is preferred by most dense backbones, while the graph backbones prefer the lighter BitFit+PA and IA\(^3\)+PA. The three exceptions, VM-UNet, Depth Pro, and Depth Anything~V2, prefer LoRA without the PA, and for these the gap to LoRA+PA is 0.03\,m or less. The full per-backbone curves are given in Appendix~\ref{app:results_transfer}.

\subsection{Qualitative Results}
\label{sec:results_qualitative}

\begin{figure*}[!h]
    \centering
    \includegraphics[width=\linewidth]{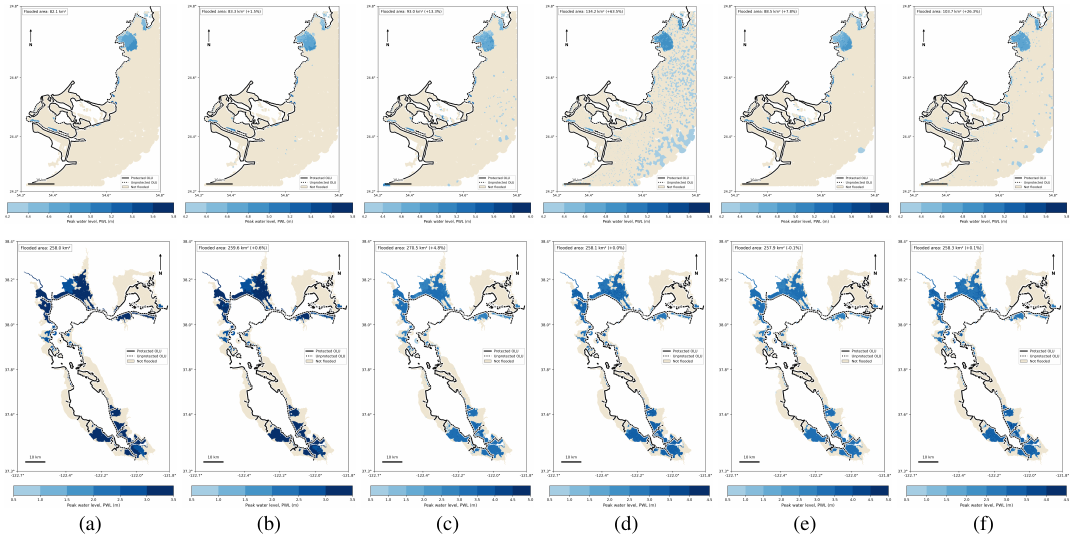}
\caption{VM-UNet PWL predictions for held-out AD (top) and SF (bottom) scenarios. (a) Ground truth, (b) in-domain, and \(K=3\) transfer using (c) FT+PA, (d) FT, (e) best PEFT+PA, and (f) best PEFT. Transfer directions are SF\(\rightarrow\)AD and AD\(\rightarrow\)SF, respectively. Each panel reports flooded area and its difference from ground truth. Color scales vary across panels.}    \label{fig:qualitative}
\end{figure*}

Figure~\ref{fig:qualitative} compares VM-UNet PWL predictions for one held-out case per region, using the in-domain model and four \(K=3\) transfer regimes. In AD, the in-domain model matches the flooded area within 1.5\%. After SF-to-AD transfer, FT and LoRA without PA overpredict flooding by 63.5\% and 26.3\%, with false wet patches over dry inland areas. Adding PA largely removes these errors, reducing the flooded-area error to 13.3\% for FT+PA and 7.8\% for PEFT+PA. This is consistent with Eq.~(\ref{eq:flood_probability}), where elevated terrain remains dry unless the learned features support flooding. It also shows that lower RMSE does not always imply a better flood extent. In SF, all AD-to-SF regimes recover the flooded area within 5\%, with differences mainly in predicted water levels inside flooded regions. Error maps are provided in Appendix~\ref{app:results_qualitative}.

\section{Conclusion}
\label{sec:conclusion}

We introduced the Physics Adapter, a small module that conditions coastal-flood predictions on terrain elevation through a differentiable wet/dry response and a gated physics-guided branch. The same formulation was attached to twelve backbones from graph, mesh, operator, vision, state-space, depth-foundation, and diffusion families and tested on geographic and SLR transfer with up to 10 labeled target scenarios. With at least one target scenario, every regime that includes the PA outperforms every regime without it in our aggregate comparison. Training the adapter alone is already better than fully fine-tuning a backbone without it, and adding it on top of LoRA, BitFit, or IA\(^3\) improves each of them. The PA also keeps or improves in-domain accuracy for ten of the twelve backbones.

\paragraph{Limitations.}
This study covers two regions, and SLR transfer is tested only within San Francisco, so broader claims need more regions and hazard types. The PA encodes a terrain comparison and not hydrodynamics, so it gives no guarantee of mass or momentum conservation. Without target data, PA source models transfer worse than raw ones, which limits zero-shot use. Finally, the interfaces, batch sizes, and PEFT insertion sites differ across architectures by design, so small differences between backbones should be read with care.

%\paragraph{Future work.}
%A natural next step is to testing the PA on riverine and pluvial flooding, %predicting time-varying water levels instead of peaks, and adding uncertainty estimates so that the adapted surrogates can support planning decisions.

%set the reference water level from information that is available without labeled simulations, such as tide-gauge records or design water levels, which could remove the zero-shot gap. Other directions include testing the PA on riverine and pluvial flooding, predicting time-varying water levels instead of peaks, and adding uncertainty estimates so that the adapted surrogates can support planning decisions.

\newpage
\subsection*{AI use statement}

In this paper, we used generative AI tools for editing and rephrasing the text to improve grammar and readability, and for drafting and editing the source code for experiments and visualization. We have not used generative AI tools to generate synthetic data sets; help develop theoretical models or conceptual frameworks; formulate mathematical claims; provide critical ingredients for proving mathematical claims; assist in the writing of proofs; propose or refine hypotheses; design or provide feedback on research  methodology or experiments; implement methods; assist with translation; clean and reformat dataset; support qualitative and thematic data analysis; and interpret results. All AI-paraphrased or AI-edited text was reviewed and revised by the authors. We take full responsibility for the final content of this work, including all text and claims produced with the aid of generative AI.

\subsection*{Ethics statement}

This work does not involve human subjects, personal data, or privacy-sensitive information. All experiments use numerical hydrodynamic simulations and publicly available geospatial datasets, used in accordance with their respective licenses. We are not aware of any conflicts of interest or other ethical concerns associated with this work. \looseness-2

\subsection*{Reproducibility statement}

We will release the full codebase, including all scripts for data preprocessing, source training, target adaptation, evaluation, and the reproduction of every table and figure in this paper. Given the size of the codebase, which spans twelve backbones and ten adaptation configurations, we are currently consolidating and documenting it, and we will share an anonymized repository link with the reviewers during the discussion period. In the meantime, the paper provides the details needed to re-implement our method and experiments.

%\subsubsection*{Author Contributions}
%If you'd like to, you may include  a section for author contributions as is done
%in many journals. This is optional and at the discretion of the authors.

%\subsubsection*{Acknowledgments}
%Use unnumbered third level headings for the acknowledgments. All acknowledgments, including those to funding agencies, go at the end of the paper.

\bibliography{iclr2027_conference}
\bibliographystyle{iclr2027_conference}

\appendix

\section*{Appendix}

% ============================================================
% A. DATA CONSTRUCTION
% ============================================================
\section{Data Construction}
\label{app:data_construction}

\begin{figure}[htb]
    \centering
    \includegraphics[width=\linewidth]{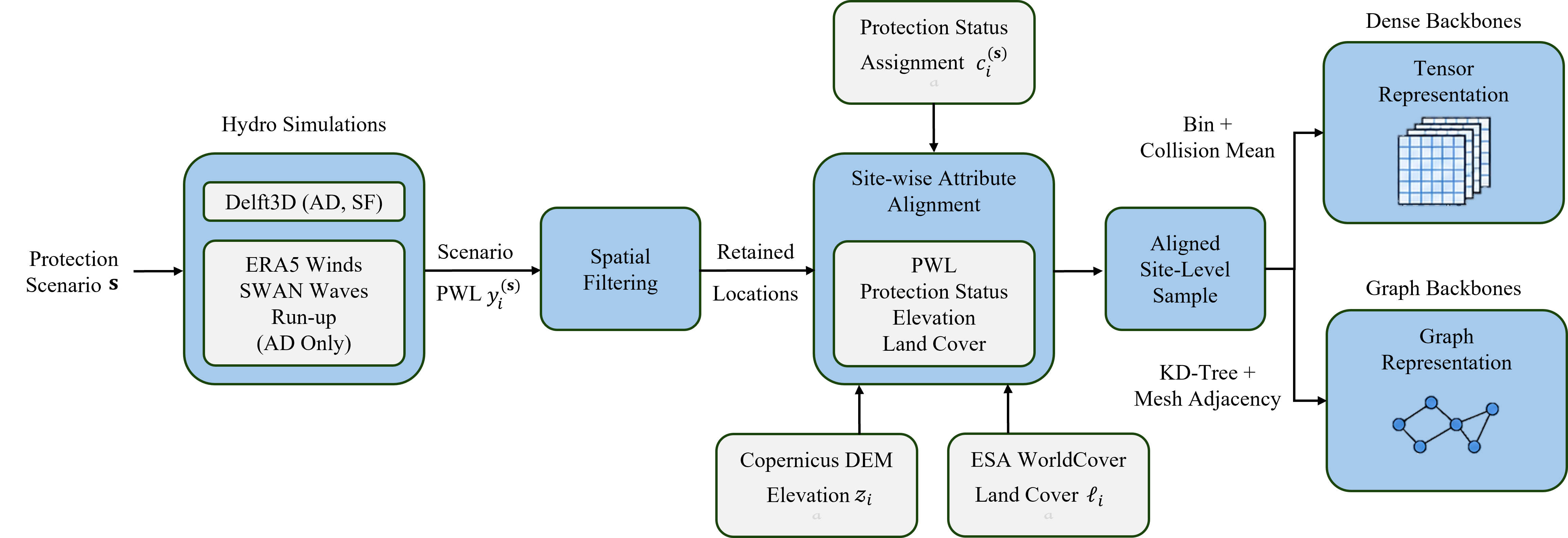}
    \caption{From hydrodynamic simulation to the two learning representations. Each retained coastal location carries scenario-dependent OLU status, elevation, and land cover, with simulated PWL as the target. Dense backbones use the rasterized tensor, and graph backbones use the mesh-topology graph built over the same locations.}
    \label{fig:data_pipeline}
\end{figure}

The datasets used in this research were built from physics-based coastal flood simulations over a common set of retained spatial locations. We filtered raw hydrodynamic outputs to the learning locations, and linked each location to its hydrodynamically derived shoreline-protection dependence, terrain elevation, land cover, and scenario-specific PWL. The resulting coordinate-level data were then encoded in two forms, a regular \(1024\times1024\) spatial tensor for dense-grid backbones and a graph that keeps the neighborhood structure of the hydrodynamic computational grid for graph-native backbones. These are two representations of the same flood-prediction problem, not separately generated datasets. Figure~\ref{fig:data_pipeline} summarizes the pipeline.

\subsection{Hydrodynamic Simulation Data}
\label{app:data_hydrodynamics}

The ground-truth flood fields come from the hydrodynamic models described in the CASPIAN studies and their supplementary material \citep{hassan2026hess}. In both regions, Delft3D \citep{lesser2004delft3d} was used to resolve time-varying coastal water levels over the computational domain under prescribed SLR, tidal forcing, shoreline-protection configurations, and the other regional forcings of the original setup. The simulator produces spatially resolved water-level time series, from which peak water level (PWL) is kept as the regression target.

The Abu Dhabi configuration also accounts for the wind and wave environment of the Arabian Gulf. The validated Delft3D model was forced with ERA5 winds \citep{hersbach2020era5}, and its results were coupled to the SWAN spectral wave model \citep{booij1999swan} to represent wind-wave generation and nearshore wave transformation. The SWAN significant wave heights were then combined with local shoreline slope to estimate coastal run-up under conditions typical of prolonged Shamal events \citep{chow2022abudhabi}. San Francisco Bay is treated differently because its shoreline lies inside a sheltered bay. The CASPIAN study did not apply SWAN there, and Delft3D alone was used to generate the SLR-driven flood fields \citep{sun2020multimodal}.

\begin{figure}[t]
    \centering
    \includegraphics[width=\linewidth]{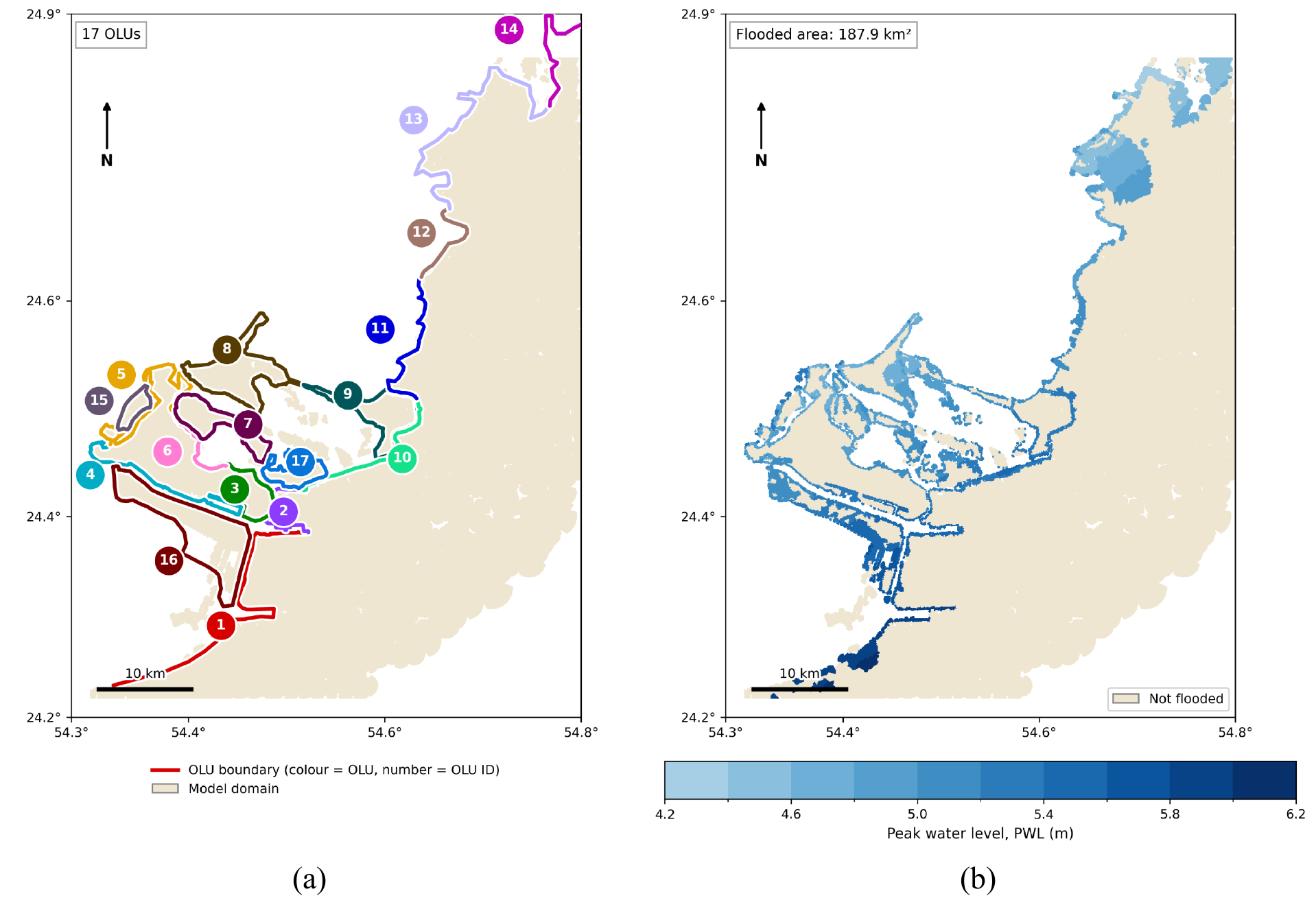}
    \caption{Abu Dhabi study region. (a)~The 17 OLUs, with each shoreline segment colored and numbered by its OLU. (b)~Simulated PWL at \(0.5\,\mathrm{m}\) SLR when no OLU is protected, giving a flooded area of 187.9\,km\(^2\).}
    \label{fig:ad_olus}
\end{figure}

\begin{figure}[t]
    \centering
    \includegraphics[width=\linewidth]{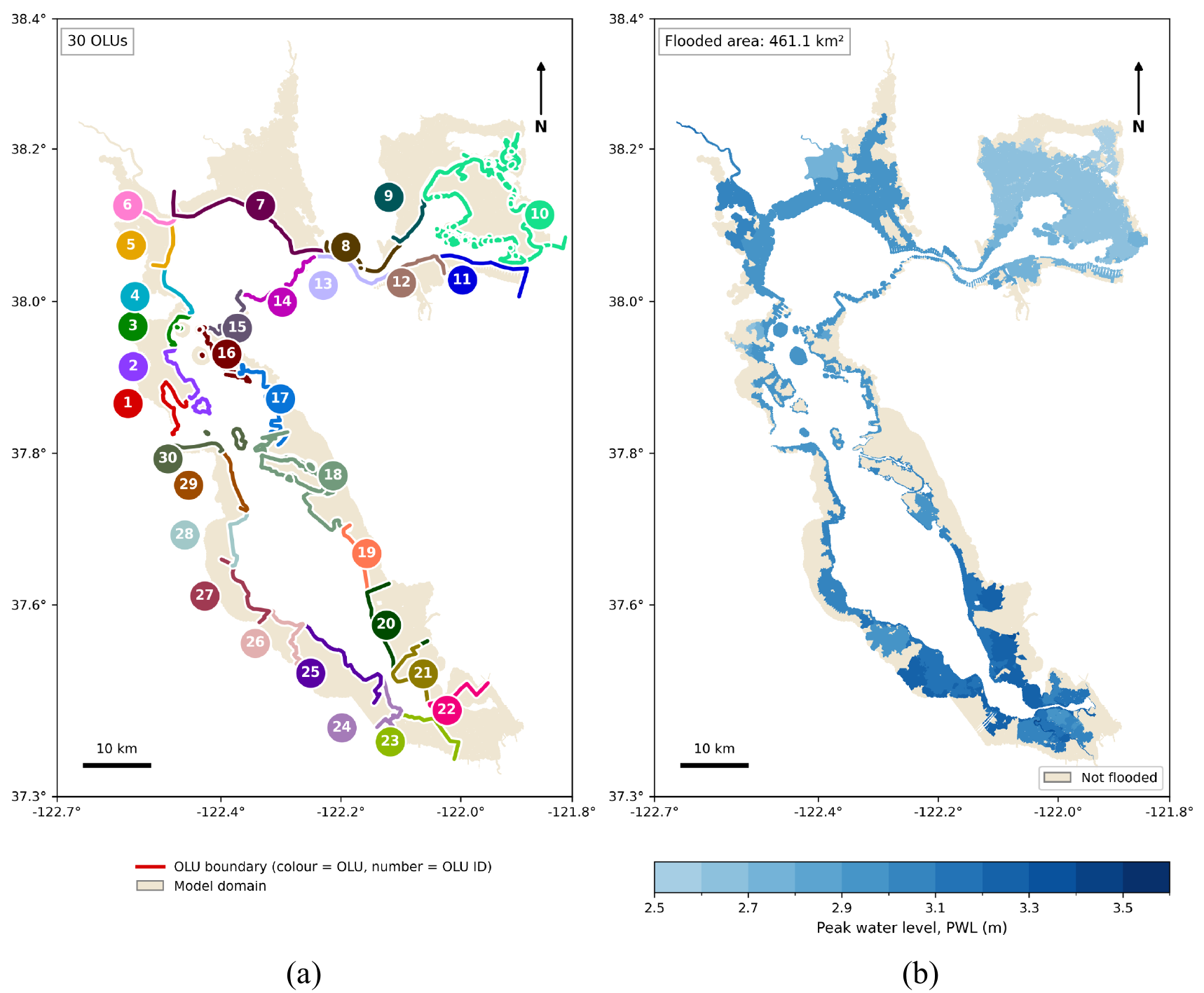}
    \caption{San Francisco Bay study region. (a)~The 30 OLUs, with each shoreline segment colored and numbered by its OLU. (b)~Simulated PWL at \(1.0\,\mathrm{m}\) SLR when no OLU is protected, giving a flooded area of 461.1\,km\(^2\).}
    \label{fig:sf_olus}
\end{figure}

The shoreline of region \(r\) is divided into \(N_r\) operational landscape units (OLUs), with \(N_{\mathrm{AD}}=17\) and \(N_{\mathrm{SF}}=30\). A protection configuration is a binary vector \(\mathbf{s}=(s_1,\ldots,s_{N_r})\), with \(s_k=1\) when OLU \(k\) is protected and \(s_k=0\) otherwise, and it is realized in the hydrodynamic model through the corresponding shoreline-defense setup. Figures~\ref{fig:ad_olus} and~\ref{fig:sf_olus} show the OLUs of each region and the flooding produced when none of them is protected. For scenario \(\mathbf{s}\), the simulator output used for learning is the set
\begin{equation}
\mathcal{R}^{(\mathbf{s})}
=
\bigl\{\bigl(\mathbf{r}_i,\,y_i^{(\mathbf{s})}\bigr)\bigr\}_{i=1}^{N},
\qquad
\mathbf{r}_i=(x_i,y_i^{\mathrm{coord}}),
\label{eq:data_raw}
\end{equation}
where \(\mathbf{r}_i\) is a simulator location and \(y_i^{(\mathbf{s})}\) its PWL. Small negative PWL values in intermediate files are clipped to zero during preprocessing. The terrain and bathymetry used inside Delft3D belong to the hydrodynamic model and are separate from the elevation feature of Appendix~\ref{app:data_auxiliary}, which is sampled independently for the learning representation.

\paragraph{Spatial curation.}
The full hydrodynamic domain contains locations that are not prediction sites for the learning task. Simulator coordinates were therefore curated to retain study-relevant coastal locations and exclude offshore, open-water, and other non-target parts of the domain. The representation scripts start from the resulting region-specific master coordinate sets and match each scenario to them. PWL, elevation, land cover, and OLU dependence are aligned on these retained locations by explicit coordinate matching rather than row order, and duplicate coordinates are removed during feature extraction. Curation does not remove persistently wet locations, since the dependence construction keeps and labels locations that stay flooded even under full protection.

\subsection{Hydrodynamically Derived OLU Dependence}
\label{app:data_olu_dependence}

\begin{figure*}[t]
    \centering
    \includegraphics[width=\linewidth]{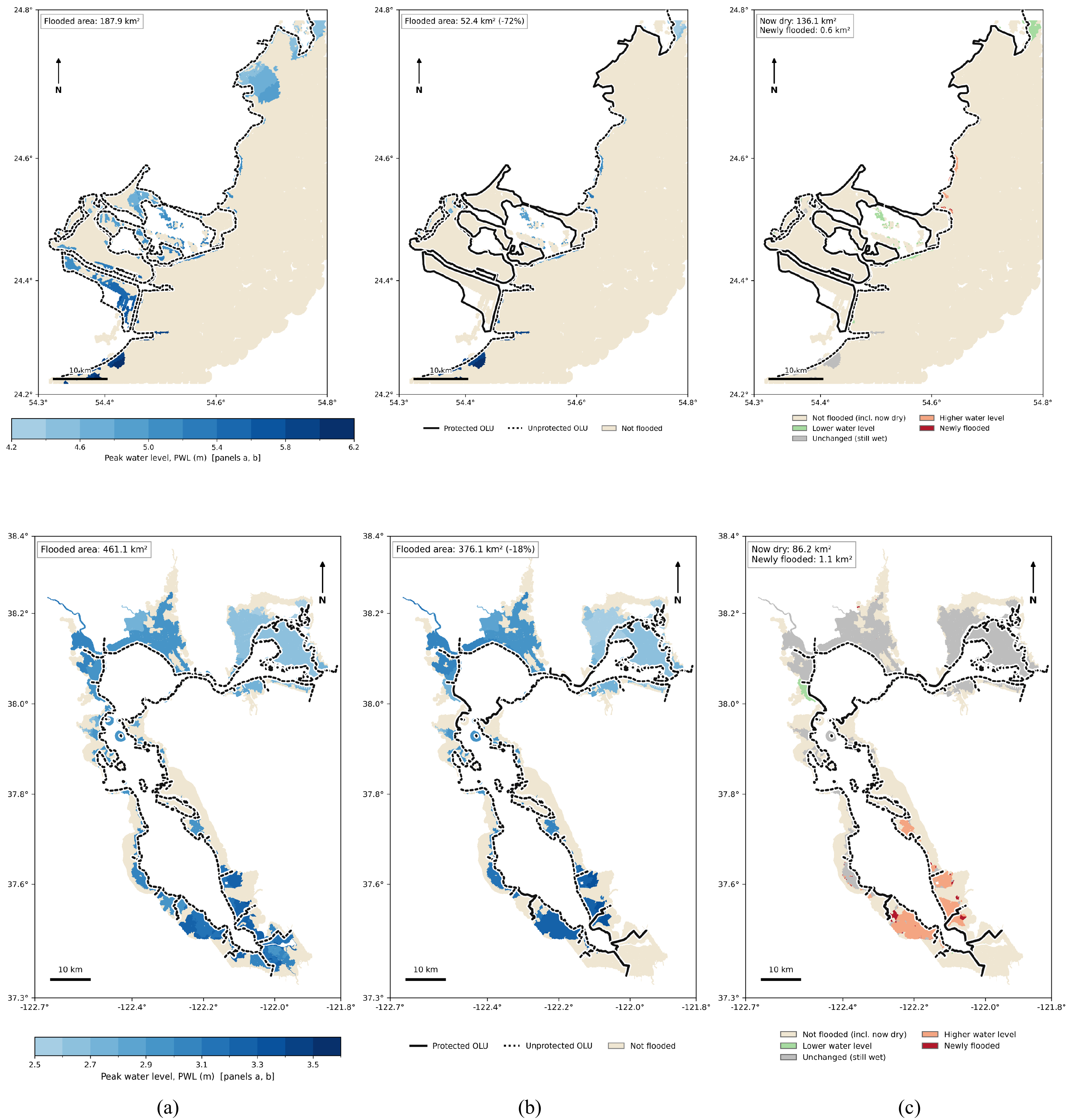}
    \caption{Effect of shoreline protection on flooding in AD (top, \(0.5\,\mathrm{m}\) SLR) and SF (bottom, \(1.0\,\mathrm{m}\) SLR). (a)~PWL when no OLU is protected. (b)~PWL for one protection scenario, with protected OLUs drawn as solid lines and unprotected OLUs as dotted lines. The flooded area falls by 72\% in AD and 18\% in SF. (c)~Change between (a) and (b). Most of the change is locations that become dry (136.1\,km\(^2\) in AD and 86.2\,km\(^2\) in SF), but some locations see a higher water level and a small area becomes newly flooded (0.6\,km\(^2\) in AD and 1.1\,km\(^2\) in SF).}
    \label{fig:olu_effect}
\end{figure*}

The protection status of a location is derived from its simulated response to OLU perturbations, not from its distance to the nearest protected or unprotected shoreline segment. We compute the dependence once per retained location and then combine it with each scenario's protection vector. Figure~\ref{fig:olu_effect} shows why proximity alone is not enough. Protecting part of the shoreline dries most of the areas behind it, but it can also raise water levels or cause new flooding elsewhere.

Let \(\mathbf{0}\) denote the all-unprotected configuration, \(\mathbf{1}\) the all-protected configuration, \(\mathbf{e}_k\) the configuration protecting only OLU \(k\), and \(\mathbf{1}-\mathbf{e}_k\) the configuration unprotecting only OLU \(k\). For every available single-OLU perturbation,
\begin{equation}
\Delta^{+}_{ik}
=
\bigl[y_i^{(\mathbf{0})}-y_i^{(\mathbf{e}_k)}\bigr]_{+},
\qquad
\Delta^{-}_{ik}
=
\bigl[y_i^{(\mathbf{1}-\mathbf{e}_k)}-y_i^{(\mathbf{1})}\bigr]_{+},
\qquad
\Delta_{ik}=\max\bigl(\Delta^{+}_{ik},\Delta^{-}_{ik}\bigr).
\label{eq:data_impact}
\end{equation}
The first term measures the PWL reduction from protecting OLU \(k\) on an otherwise unprotected shoreline, and the second measures the PWL increase from removing protection at \(k\) on an otherwise fully protected shoreline. Either direction is enough to identify an influence, and a term stays zero when its perturbation simulation is unavailable. With the largest local response \(\Delta_i^{\max}=\max_k\Delta_{ik}\), the location-specific guardian threshold is
\begin{equation}
T_i=\max\bigl(0.10~\mathrm{m},\;0.5\,\Delta_i^{\max}\bigr),
\label{eq:data_guardian_threshold}
\end{equation}
and OLU \(k\) is a guardian of location \(i\) when \(\Delta_{ik}>T_i\). Because the threshold is relative to the strongest local response, several OLUs can guard the same location. The guardian set is stored as the integer bitmask
\begin{equation}
G_i=\sum_{k=1}^{N_r}\mathbb{I}\bigl[\Delta_{ik}>T_i\bigr]\,2^{k-1},
\label{eq:data_guardian_mask}
\end{equation}
where bit \(k-1\) marks dependence on OLU \(k\). The number of guardians and the maximum-impact OLU are also recorded for diagnostics, but the bitmask is the dependence representation used downstream.

Two special cases are handled with \(\epsilon_{\mathrm{dry}}=0.05\)~m before the bitmask is finalized. A point with \(y_i^{(\mathbf{0})}<\epsilon_{\mathrm{dry}}\) stays dry even with no protection, so its guardian set is forced empty to avoid spurious dependence from numerical noise. A point with \(y_i^{(\mathbf{1})}>\epsilon_{\mathrm{dry}}\) stays wet even under full protection. Its bitmask is also cleared, and a separate always-flooded flag keeps the distinction.

\paragraph{Scenario-specific OLU status.}
With guardian set \(\mathcal{G}_i=\{k:\Delta_{ik}>T_i\}\) and unprotected OLUs \(\mathcal{U}(\mathbf{s})=\{k:s_k=0\}\), the categorical status used by both representations is
\begin{equation}
c_i^{(\mathbf{s})}
=
\begin{cases}
2, & \text{point }i\text{ is always flooded},\\
0, & \mathcal{G}_i=\varnothing,\\
2, & \mathcal{G}_i\cap\mathcal{U}(\mathbf{s})\neq\varnothing,\\
1, & \text{otherwise}.
\end{cases}
\label{eq:data_status}
\end{equation}
Status 0 means no active OLU dependence, status 1 an OLU-dependent location whose guardians are all protected, and status 2 an OLU-dependent location with at least one unprotected guardian, or an always-flooded location. The grid and graph generators use the same definition for both regions.

\subsection{Terrain and Land-Cover Attributes}
\label{app:data_auxiliary}

Each retained location is given an elevation and a land-cover value, both sampled independently of the simulator's internal bathymetry. Land cover \(\ell_i\) is sampled from ESA WorldCover 10\,m 2021 v200 \citep{zanaga2022worldcover} with its original class codes (10 tree cover, 20 shrubland, 30 grassland, 40 cropland, 50 built-up, 60 bare or sparse vegetation, 70 snow and ice, 80 permanent water bodies, 90 herbaceous wetland, 95 mangroves, and 100 moss and lichen). The grid representation keeps these codes, and the graph remaps them to contiguous categories (Appendix~\ref{app:data_graph}). Elevation \(z_i\) is sampled from Copernicus DEM GLO-30 at a nominal 30\,m resolution \citep{copernicusdem2022}. It is different from the bathymetric and terrain products used to build the Delft3D domains, and it is the elevation passed to the models and to the PA.

\begin{figure*}[t]
    \centering
    \includegraphics[width=\linewidth]{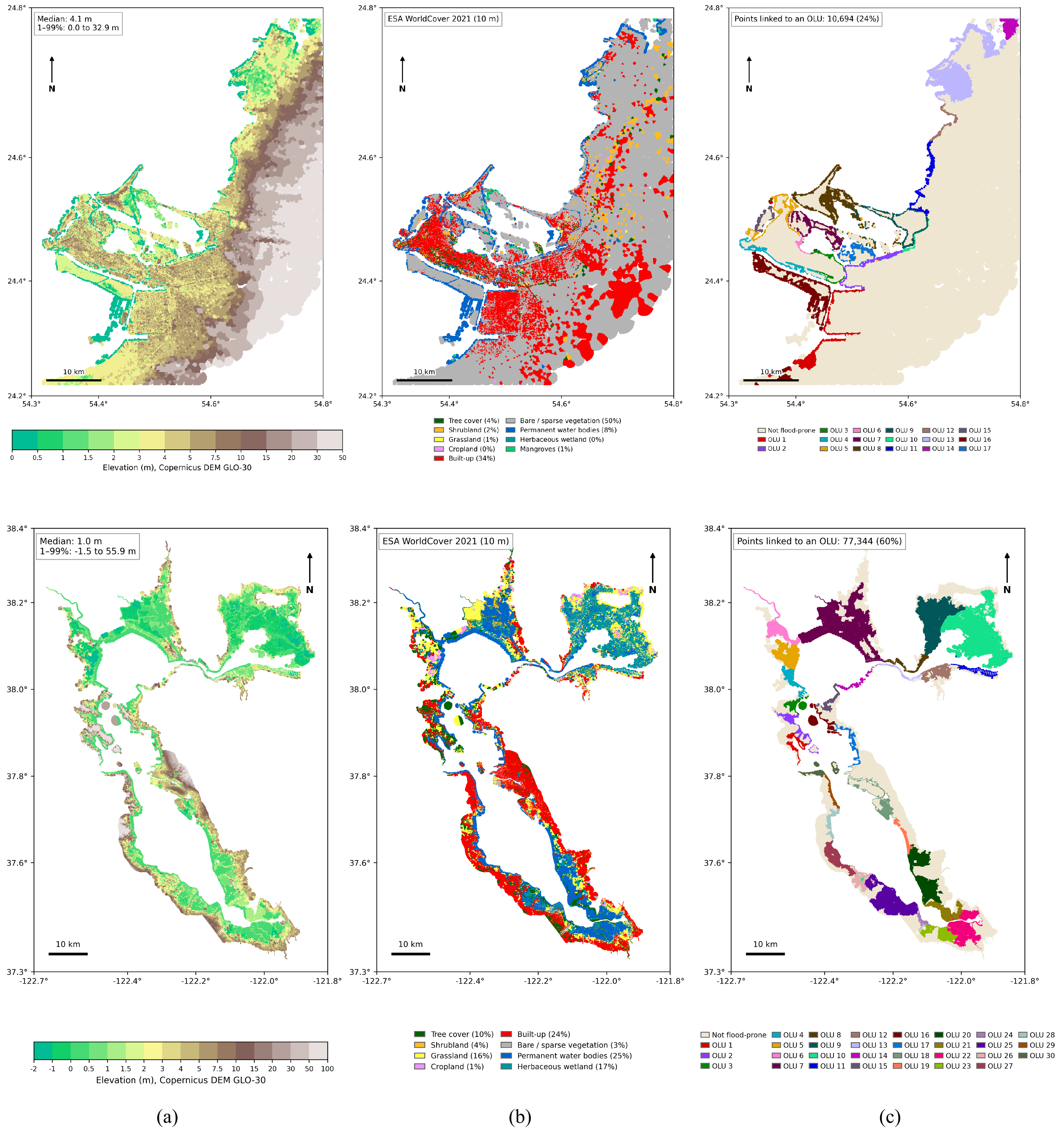}
    \caption{Input attributes for AD (top) and SF (bottom). (a)~Copernicus GLO-30 elevation, with median 4.1\,m in AD and 1.0\,m in SF. (b)~ESA WorldCover land cover. AD is dominated by bare or sparse vegetation (50\%) and built-up land (34\%), while SF has more permanent water (25\%), built-up land (24\%), and herbaceous wetland (17\%). (c)~Locations that depend on at least one OLU, colored by the OLU with the largest local impact, covering 24\% of the retained locations in AD and 60\% in SF. Locations with no OLU dependence are shown in beige. Panel (c) shows only the strongest OLU for display, while the models use the full guardian set of Eq.~(\ref{eq:data_guardian_mask}).}
    \label{fig:input_channels}
\end{figure*}

San Francisco coordinates are processed in WGS~84 / UTM Zone~10N (EPSG:32610), and Abu Dhabi coordinates in UTM Zone~40N (EPSG:32640). Both are transformed to WGS~84 geographic coordinates (EPSG:4326) before raster sampling, and land cover and elevation are read directly at the transformed coordinates from region-specific WorldCover tiles and GLO-30 rasters. Each location therefore has three model attributes \((c_i^{(\mathbf{s})},z_i,\ell_i)\) and the target PWL \(y_i^{(\mathbf{s})}\). The occupancy mask introduced below is a structural grid indicator, not a physical attribute. Figure~\ref{fig:input_channels} shows elevation, land cover, and the OLU dependence for both regions.

\subsection{Regular-Grid Representation}
\label{app:data_grid}

Dense backbones use a \(1024\times1024\) representation of the retained coordinates. The generator uses natural geographic bins with \(N_g=1024\) and does not move colliding points into neighboring cells. Let \(x_{\min},x_{\max},y_{\min},y_{\max}\) be the extrema of the retained coordinates. A \(2\%\) margin of the coordinate range is added on each axis, giving \(\widetilde{x}_{\min},\widetilde{x}_{\max},\widetilde{y}_{\min},\widetilde{y}_{\max}\), and each coordinate is assigned the bin
\begin{equation}
u_i
=
\operatorname{clip}\!\left(
\left\lfloor N_g\,\tfrac{x_i-\widetilde{x}_{\min}}{\widetilde{x}_{\max}-\widetilde{x}_{\min}}\right\rfloor,
0,N_g{-}1\right),
\qquad
w_i
=
\operatorname{clip}\!\left(
\left\lfloor N_g\,\tfrac{y_i^{\mathrm{coord}}-\widetilde{y}_{\min}}{\widetilde{y}_{\max}-\widetilde{y}_{\min}}\right\rfloor,
0,N_g{-}1\right).
\label{eq:data_grid_bins}
\end{equation}
This mapping is deterministic and stored for every retained coordinate.

Because the hydrodynamic locations are irregularly spaced, several coordinates can fall into the same bin, and they stay there. For each occupied bin, the coordinates are sorted lexicographically and the first \((x,y)\) pair supplies the status, elevation, and land-cover inputs. The target instead uses all points in the bin. With \(\mathcal{C}_{uw}=\{i:(u_i,w_i)=(u,w)\}\),
\begin{equation}
Y_{uw}^{(\mathbf{s})}
=
\frac{1}{|\mathcal{C}_{uw}|}\sum_{i\in\mathcal{C}_{uw}} y_i^{(\mathbf{s})},
\qquad
M_{uw}=\mathbb{I}\bigl[|\mathcal{C}_{uw}|>0\bigr],
\label{eq:data_conflict_mean}
\end{equation}
so a single point keeps its own PWL and a shared bin receives the mean. The occupancy mask \(M_{uw}\) fills the validity channel \(\mathbf{v}\) of Sec.~\ref{sec:problem} and separates occupied prediction sites from empty background. Validity cannot be inferred safely from a zero DEM, land-cover, or PWL value.

For scenario \(\mathbf{s}\), the stored input tensor has shape \(1024\times1024\times4\) with channel order OLU status, DEM, WorldCover class, and occupancy mask. The target is a \(1024\times1024\) matrix with non-negative PWL at occupied bins and zero elsewhere. Both arrays are built with the \((u,w)\) indexing of Eq.~(\ref{eq:data_grid_bins}), and their two spatial axes are transposed together before saving, so they share the same orientation. The stored geographic mapping keeps the full coordinate-to-bin map and the coordinates of every occupied bin. A reconstruction utility uses it to return grid predictions to the coordinate level, where coordinates sharing a bin take that bin's prediction.

\subsection{Graph Representation}
\label{app:data_graph}

Graph-native backbones use the same retained locations and scenario definitions but keep the neighborhood structure of the hydrodynamic grid instead of rasterizing. Each scenario graph has one node per retained coordinate. Node \(i\) stores the continuous feature \([z_i]\), the categorical features \([\widetilde{\ell}_i,\,c_i^{(\mathbf{s})}]\) as integer indices, the position \([x_i,y_i^{\mathrm{coord}}]\), and the target \(y_i^{(\mathbf{s})}\). WorldCover codes are remapped to contiguous categories \(10\!\rightarrow\!0\), \(20\!\rightarrow\!1\), \(30\!\rightarrow\!2\), \(40\!\rightarrow\!3\), \(50\!\rightarrow\!4\), \(60\!\rightarrow\!5\), \(70\!\rightarrow\!6\), \(80\!\rightarrow\!7\), \(90\!\rightarrow\!8\), \(95\!\rightarrow\!9\), and \(100\!\rightarrow\!10\), with unknown values assigned category 11. The OLU status is the same variable as in Eq.~(\ref{eq:data_status}). Unlike the grid, the graph contains only prediction nodes and needs no occupancy channel.

\paragraph{Hydrodynamic-grid connectivity.}
Edges follow the topology of the hydrodynamic computational grid rather than a generic \(k\)-nearest-neighbor rule. The generator reads cell centers and face-node coordinates from the region-specific grid files and links each retained coordinate to its nearest cell center with a KD-tree. The San Francisco grid is already in UTM Zone~10N. The Abu Dhabi grid is stored in EPSG:4326, so its cell centers and face nodes are transformed to EPSG:32640 before matching. Two cells are adjacent when they share a complete boundary edge, each adjacent pair is added in both directions, and the adjacency is then restricted to the cells linked to retained coordinates. For an edge \(i\rightarrow j\), the edge feature is \(\mathbf{e}_{ij}=[\Delta x_{ij},\Delta y_{ij},d_{ij}]\), with \(\Delta x_{ij}=x_i-x_j\), \(\Delta y_{ij}=y_i^{\mathrm{coord}}-y_j^{\mathrm{coord}}\), and Euclidean distance \(d_{ij}\). No separate simulations are run for the graph. Scenario PWL, DEM, and land cover are aligned by coordinate before construction, and the same guardian logic gives the scenario-dependent status.

\paragraph{Consistency across representations.}
In both encodings, a sample is defined by the same retained coordinates, OLU configuration, guardian dependence, Copernicus elevation, WorldCover class, and PWL target. The grid merges points only when they share a geographic bin, while the graph keeps every point as a separate node linked by mesh topology. Any later normalization, projection, or embedding belongs to the model, not to the dataset.

% ============================================================
% B. ARCHITECTURE INTERFACES
% ============================================================
\section{Architecture-Specific Physics Adapter Interfaces}
\label{app:architecture_interfaces}

Table~\ref{tab:backbone_interfaces} lists, for each backbone, the representation passed to the PA and how the interface \(\mathcal{P}_m\) of Eq.~(\ref{eq:pipeline}) is realized.

\begin{table*}[t]
\centering
\small
\setlength{\tabcolsep}{4pt}
\begin{tabular}{p{0.14\textwidth} p{0.17\textwidth} p{0.28\textwidth} p{0.32\textwidth}}
\hline
\textbf{Family} &
\textbf{Backbone} &
\textbf{Representation passed to PA} &
\textbf{Interface realization} \\
\hline
Graph &
GCN \citep{kipf2017gcn} &
Node-aligned latent embeddings &
Node-wise PA heads over retained graph locations. \\
Graph attention &
GAT \citep{velickovic2018gat} &
Node-aligned attention embeddings &
Node-wise PA heads over retained graph locations. \\
Mesh graph &
MGN \citep{pfaff2021meshgraphnets} &
Node-aligned latent embeddings &
Node-wise PA heads after native mesh message passing. \\
Scientific attention &
Transolver++ \citep{luo2025transolverpp} &
Site-aligned operator representation &
Node-wise PA heads at the physical prediction sites. \\
\hline
Dense vision &
CASPIAN \citep{karapetyan2026caspian} &
Dense spatial feature map &
Dense features, then pointwise PA heads. \\
Dense vision &
ConvNeXt~V2 \citep{woo2023convnextv2} &
Dense and multi-scale visual features &
Dense decoding and alignment, then pointwise PA heads. \\
Dense vision &
MaxViT \citep{tu2022maxvit} &
Dense and multi-scale visual features &
Dense decoding and alignment, then pointwise PA heads. \\
Dense vision &
Swin~V2 \citep{liu2022swinv2} &
Dense and multi-scale transformer features &
Dense decoding and alignment, then pointwise PA heads. \\
\hline
State space &
VM-UNet \citep{ruan2024vmunet,liu2024vmamba} &
Decoder feature before the native final regression layer &
Decoder feature projected to the PA width, then pointwise PA heads. \\
Depth foundation &
Depth Anything~V2 \citep{yang2024depthanythingv2} &
Pretrained dense DPT feature before depth regression &
Flood-specific input stem and feature projection align the feature to the PWL grid. \\
Depth foundation &
Depth Pro \citep{bochkovskiy2025depthpro} &
Pretrained dense depth representation &
Task-specific projection and interpolation align the feature to the PWL grid. \\
Diffusion &
ControlNet \citep{zhang2023controlnet} &
Generated base PWL map with local conditioning &
Adapter-owned post-decoder stem, with the data branch predicting a residual around the generated map. \\
\hline
\end{tabular}
\caption{Backbones and their realization of the shared PA interface. The representation column describes the site-aligned quantity passed to the PA, not the full internal architecture. Terrain conditioning, the two prediction branches, gated fusion, and the adaptation protocol are shared across all rows.}
\label{tab:backbone_interfaces}
\end{table*}

\paragraph{Adapter normalization and heads.}
For raster representations, \(\operatorname{BN}^{(m)}_{A}\) in Eq.~(\ref{eq:adapter_fields}) is a channel-wise \(\operatorname{BatchNorm2d}\), and node-aligned implementations use \(\operatorname{BatchNorm1d}\). The CASPIAN, ConvNeXt~V2, GAT, GCN, MGN, and Transolver++ adapters have affine BatchNorm parameters, while the MaxViT, Swin~V2, VM-UNet, Depth Anything~V2, Depth Pro, and ControlNet adapters use affine-free BatchNorm. The data branch applies dropout before its final prediction, with rate \(0.10\) in the dense and diffusion adapters and \(0.30\) in the node-aligned adapters.

\paragraph{Initialization.}
The temperature is initialized to \(\tau=0.5\), and \(\eta\) is initialized from the source-region reference level, \(3.0\) for San Francisco and \(5.0\) for Abu Dhabi. Both \(\eta\) and \(\vartheta\) are then learned. The \(+3\) gate initialization of Eq.~(\ref{eq:fusion}) is a fixed additive logit offset in the dense and ControlNet adapters and an initial value of the final gate-head bias in the node-aligned adapters. This changes the parameterization but not the meaning of Eq.~(\ref{eq:fusion}). In dense implementations, the gate penalty of Eq.~(\ref{eq:objective}) is computed over the full gate tensor before validity masking, so \(\Omega_A\) covers all grid cells. In node-aligned implementations, it is computed over the represented nodes.

\paragraph{State-space and depth-foundation interfaces.}
In VM-UNet, the representation just before the native final regression layer is projected to the adapter width before entering the adapter BatchNorm. Depth Anything~V2 similarly takes a dense DPT feature before the original depth regressor, using a learned flood-input stem and a feature projection. In both cases, the raw DEM bypasses this path and enters Eq.~(\ref{eq:flood_probability}) directly. These operations belong to the interface, not to the shared adapter equations.

\paragraph{ControlNet interface and objective.}
The output of ControlNet is generative rather than a deterministic regression field, so its interface differs from the other backbones. The pretrained SD3.5 transformer and VAE stay frozen by design. The trainable ControlNet branch produces the conditional representation, and a clean-latent estimate is decoded to a base PWL map. A small adapter-owned convolutional stem then combines this map with protection status, DEM, and land cover before the shared adapter computation, so the stem parameters belong to \(\phi_m\). The data branch predicts a residual around the generated base value \(b_i\). ControlNet keeps its diffusion training objective alongside the map objective. Writing \(\mathbf{l}_0\) for a clean diffusion latent, \(\widehat{\mathbf{l}}_0\) for the preconditioned clean-latent estimate, and \(\varsigma\) for the noise level,
\begin{equation}
\mathcal{L}_{\mathrm{flow}}
=
\mathbb{E}\bigl[w(\varsigma)\,\bigl\|\widehat{\mathbf{l}}_0-\mathbf{l}_0\bigr\|_2^2\bigr],
\qquad
\mathcal{L}_{\mathrm{ControlNet}}
=
\mathbb{I}_{\theta_{\mathrm{CN}}}\mathcal{L}_{\mathrm{flow}}
+
\mathbb{I}_{\mathrm{route}}\bigl(\mathcal{L}_{\mathrm{pred}}+\mathbb{I}_{\mathrm{PA}}\mathcal{L}_{g}\bigr),
\label{eq:controlnet_loss}
\end{equation}
where the indicators depend on the adaptation regime. The decoded base map is detached from the diffusion computation. Map-level gradients therefore train the post-decoder PA or raw head, and diffusion gradients train the ControlNet branch when it is trainable. For ControlNet, \(\theta_m\) denotes the trainable ControlNet branch and its interface.

\paragraph{Raw-head regularization.}
The raw head \(\chi_m\) is regularized independently of the PA. Raster backbones use a statistics-free GroupNorm and convolutional prediction head, and graph backbones use a node-wise normalized MLP. Both use train-time feature jitter and dropout, with a fixed weight decay of \(10^{-4}\) on the head weights. This makes the raw model a regularized no-PA control rather than a plain linear probe.

Because the PA realization differs across families in BatchNorm affine state, convolutional or MLP heads, and the ControlNet stem, its trainable parameter count is measured separately for each backbone.

% ============================================================
% C. ADAPTATION AND PEFT
% ============================================================
\section{Adaptation and PEFT Details}
\label{app:adaptation_details}

\begin{figure*}[t]
    \centering
    \includegraphics[width=\linewidth]{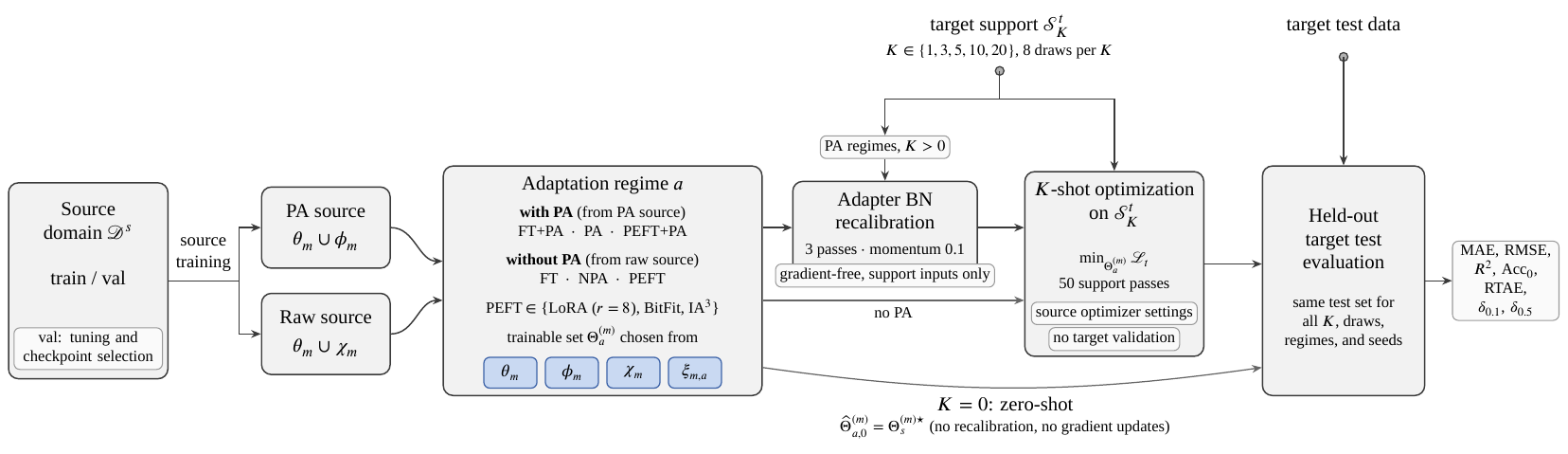}
    \caption{Two-stage training and adaptation protocol. Source checkpoints are selected with source-domain validation only. Target support examples may update only the state permitted by the selected regime, and target-test examples are held out until final evaluation; at \(K=0\), both support-statistics recalibration and gradient adaptation are skipped.}
    \label{fig:adaptation_protocol}
\end{figure*}

Table~\ref{tab:adaptation_modes_full} lists the ten target-adaptation regimes of Sec.~\ref{sec:benchmarks}, and Figure~\ref{fig:adaptation_protocol} summarizes the two-stage protocol.

\begin{table*}[t]
\centering
\small
\setlength{\tabcolsep}{4pt}
\begin{tabular}{p{0.15\textwidth} p{0.14\textwidth} p{0.31\textwidth} p{0.31\textwidth}}
\hline
\textbf{Regime} &
\textbf{Source checkpoint} &
\textbf{Trainable during adaptation} &
\textbf{Frozen during adaptation} \\
\hline
FT+PA &
PA source &
Trainable backbone and interface, and PA parameters \(\phi_m\) &
Only components fixed by design (e.g., the SD3.5 foundation in ControlNet). \\
FT &
Raw source &
Trainable backbone and interface, and prediction head \(\chi_m\) &
Only components fixed by design (e.g., the SD3.5 foundation in ControlNet). \\
NPA &
Raw source &
Prediction head \(\chi_m\) &
Backbone and interface. \\
PA &
PA source &
PA parameters \(\phi_m\) &
Backbone and interface. \\
\hline
LoRA (PEFT) &
Raw source &
Rank-8 LoRA parameters &
Prediction head and all base parameters. \\
BitFit (PEFT) &
Raw source &
Selected bias parameters &
Prediction head and all other parameters. \\
IA\(^3\) (PEFT) &
Raw source &
IA\(^3\) scaling parameters &
Prediction head and all base parameters. \\
\hline
LoRA+PA (PEFT+PA) &
PA source &
\(\phi_m\) and rank-8 LoRA parameters &
All base backbone and interface parameters. \\
PA+BitFit (PEFT+PA) &
PA source &
\(\phi_m\) and selected bias parameters &
All other backbone and interface parameters. \\
PA+IA\(^3\) (PEFT+PA) &
PA source &
\(\phi_m\) and IA\(^3\) scaling parameters &
All base backbone and interface parameters. \\
\hline
\end{tabular}
\caption{The ten target-adaptation regimes. FT+PA and FT denote full fine-tuning with and without the PA, and NPA denotes partial fine-tuning of the prediction head \(\chi_m\) without the PA. In the PEFT regimes without the PA, the prediction head stays frozen, so only the PEFT parameters are updated. PEFT parameters are injected after the source checkpoint is loaded and attach to architecture-compatible modules, so their locations and counts differ across backbones.}
\label{tab:adaptation_modes_full}
\end{table*}

% ============================================================
% FIGURE PLACEHOLDER A2 --- replace with final figure
% ============================================================

\paragraph{PEFT formulations.}
LoRA \citep{hu2022lora} replaces a selected frozen linear or channel-mixing transformation \(\mathbf{W}_0\) by
\begin{equation}
\mathbf{W}\mathbf{x}
=
\mathbf{W}_0\mathbf{x}+\frac{\alpha}{r}\,\mathbf{B}\mathbf{A}\mathbf{x},
\qquad r=8,\quad \alpha=16,
\label{eq:lora}
\end{equation}
with \(\mathbf{A}\) Kaiming-initialized and \(\mathbf{B}=0\), so the injected update starts at zero. IA\(^3\) \citep{liu2022ia3} learns multiplicative activation scales \(\mathbf{y}=\mathbf{s}_{\mathrm{IA}}\odot(\mathbf{W}_0\mathbf{x})\), initialized at \(\mathbf{1}\). BitFit \citep{benzaken2022bitfit} updates selected bias parameters and keeps all other tensors fixed.

\paragraph{Insertion sites.}
PEFT modules are placed according to each backbone's structure. They target attention and feed-forward projections in attention-based backbones, channel-mixing layers in convolutional models, and the SS2D input and output projections in VM-UNet, where the selective-scan recurrence tensors are not LoRA or IA\(^3\) targets. In graph, operator, and ControlNet modules, they target compatible linear layers. In the graph and operator implementations, PEFT injection runs recursively over all compatible linear layers and can therefore also place PEFT parameters inside node-wise head modules. These parameters still belong to \(\xi_{m,a}\), and the base parameters still follow Table~\ref{tab:adaptation_modes_full}. In the PEFT+PA regimes, \(\phi_m\) and \(\xi_{m,a}\) are optimized jointly. The trainable-parameter fraction of regime \(a\) is
\begin{equation}
\rho_a^{(m)}
=
\frac{|\Theta_a^{(m)}|}{|\Theta_{\mathrm{total}}^{(m)}|}\times100\%,
\label{eq:parameter_fraction}
\end{equation}
which describes the size of the optimization problem only and says nothing about performance by itself.

\paragraph{Freezing.}
A frozen component is frozen in both its parameters and its internal state. Setting \texttt{requires\_grad=False} is not enough for modules with BatchNorm, Dropout, stochastic depth, or other stateful operations. During adaptation with a frozen backbone, backbone normalization layers stay in inference mode and frozen Dropout, DropPath, and other stochastic modules are disabled. The implementation re-applies these settings after every change of training mode and checks that frozen running statistics stay unchanged. The two full fine-tuning regimes are the only exception, and there the trainable backbone updates normally.

\paragraph{Adapter BatchNorm recalibration.}
For every regime with a PA and \(K>0\), adaptation begins with a gradient-free recalibration of the adapter BatchNorm of Eq.~(\ref{eq:adapter_fields}). The whole model is set to inference mode, only the adapter BatchNorm is switched to training mode, and over three complete passes of the support set its running statistics are updated as
\begin{equation}
\widehat{\boldsymbol{\mu}}_{A}
\leftarrow
(1-\mu_{\mathrm{BN}})\,\widehat{\boldsymbol{\mu}}_{A}+\mu_{\mathrm{BN}}\,\boldsymbol{\mu}_{\mathcal{B}},
\qquad
\widehat{\boldsymbol{\sigma}}_{A}^{2}
\leftarrow
(1-\mu_{\mathrm{BN}})\,\widehat{\boldsymbol{\sigma}}_{A}^{2}+\mu_{\mathrm{BN}}\,\boldsymbol{\sigma}_{\mathcal{B}}^{2},
\qquad
\mu_{\mathrm{BN}}=0.1,
\label{eq:support_bn}
\end{equation}
with no optimizer step or gradient. The mini-batch size for this step depends on the architecture. It is four support graphs for GAT, GCN, MGN, and Transolver++, two samples for CASPIAN and ConvNeXt~V2, and one sample for MaxViT, Swin~V2, VM-UNet, Depth Anything~V2, Depth Pro, and ControlNet. For ControlNet, the base map used here is generated from support conditioning only. Neither support labels nor target test samples are used to estimate the adapter statistics. When \(\phi_m\) stays trainable during the supervised support optimization that follows, its BatchNorm keeps updating from the same support data, and any affine BatchNorm parameters receive gradients as part of \(\phi_m\). All frozen backbone statistics remain fixed. Regimes without a PA skip this step.

\paragraph{Fixed support budget.}
After recalibration, the parameters selected in Eq.~(\ref{eq:target_optimization}) are optimized on \(\mathcal{S}_{K}^{t}\) only. The eleven non-diffusion backbones run 50 complete passes over the support set, with one optimizer update per support mini-batch in each pass. The value 50 is therefore a fixed number of passes, which gives more than 50 updates when \(K\) exceeds the support batch size. ControlNet instead runs exactly 50 optimizer steps while cycling through its support loader with its configured gradient accumulation. The optimizer, learning rate, weight decay, and other settings are taken from source training, gradients are clipped to global norm 1, and no target validation, early stopping, or test-based checkpoint selection is used. Each combination of regime, \(K\), seed, and support draw starts once from its source checkpoint, is adapted once, and is then evaluated on the target test set.

% ============================================================
% D. SPLITS AND SUPPORT CONSTRUCTION
% ============================================================
\section{Scenario Splits and Support Construction}
\label{app:splits}

Table~\ref{tab:protocol} summarizes the scenario partitions and evaluation settings. Source training uses the training split, hyperparameter search, early stopping, and checkpoint selection use only the validation split, and the test split is used only for final evaluation. A PA source model and a raw source model are trained for every backbone, region, and seed.

\begin{table}[t]
\centering
\small
\caption{Scenario partitions and evaluation settings. For the regional datasets, the scenario column gives train/validation/test counts. For the SLR targets, it gives the support pool and held-out test, with no validation partition. For \(K>0\), eight support draws are evaluated per \(K\), and \(K=0\) is evaluated once per seed.}
\label{tab:protocol}
\begin{tabular}{llllr}
\hline
Experiment & Source \(\rightarrow\) target & Scenarios & \(K\) & Seeds \\
\hline
In-domain SF   & SF \(\rightarrow\) SF & 285 (168/57/60) & -- & 3 \\
In-domain AD   & AD \(\rightarrow\) AD & 142 (82/28/32) & -- & 3 \\
Cross-region   & SF \(\rightarrow\) AD & 142 (82/28/32) & 0,1,3,5,10 & 3 \\
Cross-region   & AD \(\rightarrow\) SF & 285 (168/57/60) & 0,1,3,5,10 & 3 \\
Cross-SLR      & \(\mathrm{SF}_{1.0}\rightarrow\mathrm{SF}_{0.5}\) & 32 (26 pool / 6 test) & 0,1,3,5,10 & 3 \\
Cross-SLR      & \(\mathrm{SF}_{1.0}\rightarrow\mathrm{SF}_{1.5}\) & 32 (26 pool / 6 test) & 0,1,3,5,10 & 3 \\
\hline
\end{tabular}
\end{table}

\paragraph{Protection buckets and regional splits.}
Each scenario is identified by its OLU configuration \(\mathbf{s}\), with protection level \(n_{\mathrm{prot}}(\mathbf{s})=\sum_k s_k\). The all-unprotected and all-protected configurations form their own \texttt{none} and \texttt{all} buckets. The remaining scenarios are split by the quartiles of \(n_{\mathrm{prot}}\) into \texttt{very\_low}, \texttt{low}, \texttt{mid}, and \texttt{high}, with boundaries \(Q_{25}=7\), \(Q_{50}=8\), \(Q_{75}=9\) for SF and \(Q_{25}=4\), \(Q_{50}=8\), \(Q_{75}=11\) for AD. Scenarios with \(n_{\mathrm{prot}}\leq Q_{25}\), \(Q_{25}<n_{\mathrm{prot}}\leq Q_{50}\), and \(Q_{50}<n_{\mathrm{prot}}\leq Q_{75}\) fall into the first three buckets, and the rest into \texttt{high}. The buckets therefore follow the observed scenario distribution rather than equal protection intervals.

For a protection bucket \(b\) with \(n_b\) scenarios, the number of test scenarios is
\begin{equation}
n_{\mathrm{test}}(b)=
\begin{cases}
1, & n_b=1,\\
\min\bigl\{n_b-1,\;\max\bigl(2,\lceil 0.2\,n_b\rceil\bigr)\bigr\}, & n_b>1,
\end{cases}
\label{eq:data_test_allocation}
\end{equation}
which keeps at least one non-test scenario in any bucket with more than one sample. The single-scenario \texttt{none} and \texttt{all} buckets both go to test. This gives 60 SF and 32 AD test scenarios. With the test sets fixed, the remaining scenarios are split again by the same buckets, aiming for a validation share of about \(20\%\) of the full regional dataset. The result is 168/57/60 train/validation/test scenarios for SF and 82/28/32 for AD, or roughly \(60/20/20\) after rounding within buckets. The regional manifest is generated once with seed 42 and shared by all twelve backbones and all training seeds, with no architecture-specific resampling, so differences between models cannot come from different partitions.

\paragraph{Cross-region support draws.}
Cross-region support sets are drawn from the target non-test pool, which is the union of the target train and validation partitions, so that \(\mathcal{S}^{t}_{K}\subseteq\mathcal{D}^{t}_{\mathrm{train}}\cup\mathcal{D}^{t}_{\mathrm{val}}\) and \(\mathcal{S}^{t}_{0}=\varnothing\). Eight reproducible draws are made for each \(K\in\{0,1,3,5,10,20\}\). Draws are stratified by protection bucket, with bucket shares roughly matching the non-test pool, and within each \(K\), scenarios used less often in earlier draws are preferred to limit repetition. The target test set is the same for all \(K\), so differences across \(K\) reflect the amount of support rather than the test scenarios. In both transfer directions, the source model is trained on the source training split and selected on the source validation split.

\paragraph{SLR target manifests.}
The SLR targets use their own fixed construction instead of the 285-scenario SF split. The \(0.5\,\mathrm{m}\) and \(1.5\,\mathrm{m}\) SF targets each contain 32 configurations, namely the all-unprotected and all-protected anchors and the 30 single-OLU configurations. The two anchors are never used for testing. Six single-OLU configurations, with OLU indices spread evenly over the index range, are held out as the test set (seed 42), and the remaining 24 single-OLU configurations and the two anchors form a 26-scenario support pool. No validation partition is defined. The pool is stored under the \texttt{train} field of the transfer manifest for loader compatibility, but it serves only as the support pool. SLR adaptation uses \(K\in\{0,1,3,5,10\}\) with eight draws per \(K\). All \(K=0\) draws are empty. At \(K=1\), two draws use the all-unprotected anchor, two use the all-protected anchor, and four use distinct single-OLU configurations. For \(K\geq3\), both anchors are always included and the remaining \(K-2\) positions are distinct single-OLU configurations. All \(K\) are evaluated on the same six test scenarios, and no test scenario appears in any support set.

% ============================================================
% E. TRAINING AND HPO
% ============================================================
\section{Training and HPO Details}
\label{app:training_details}

Hyperparameter optimization (HPO) searches only optimization settings, namely a log-scaled learning rate in \([10^{-5},5\times10^{-3}]\), the optimizer (Adam, AdamW, or RMSprop), and a feasible batch size. Architectural settings are never searched. %Each study runs 25 Optuna trials of at most 80 epochs, separately for each training seed. The test split is never used by HPO.

%Final source training runs for at most 400 epochs, with early stopping after 20 epochs without improvement in validation MSE. We keep the checkpoint with the lowest validation MSE. Appendix~\ref{app:training_details} provides the learning-rate schedule, gradient clipping, and other architecture-specific settings.

HPO runs 25 Optuna trials of at most 80 epochs per study and minimizes masked validation MSE. Batch-size candidates are set separately for each architecture according to what fits in memory, so no batch-size set is reused across backbones. The selected hyperparameters, optimizer settings, and batch sizes are kept fixed after HPO for all subsequent training and adaptation runs.

Source training uses \texttt{ReduceLROnPlateau} on validation loss with factor \(0.5\), patience of 5 epochs, and minimum learning rate \(10^{-6}\). Gradients are clipped to global norm 1 in both source training and target adaptation. Weight decay is applied through the optimizer and is not part of Eq.~(\ref{eq:objective}). Final source training runs for at most 400 epochs for all backbones, with early stopping after 20 epochs without improvement in validation MSE. Early stopping usually ends training well before this limit, and the checkpoint with the lowest validation MSE is kept for evaluation and transfer.

The random state used to generate the manifest is separate from the training seeds \(\{0,1,2\}\). Dense-grid models use seed-controlled spatial flips as training augmentation, with validation and test samples left unaugmented. Augmentation for the other families is architecture-specific.

% ============================================================
% F. EVALUATION
% ============================================================
\section{Evaluation and Complexity Details}
\label{app:evaluation_details}

\paragraph{Metric definitions.}
All models are scored at the same retained physical locations. Grid predictions are first mapped back to these locations through the stored geographic mapping, while graph models already predict on them. For a held-out scenario with \(N\) retained locations, true PWL \(y_i\), predicted PWL \(\widehat{y}_i\), and mean true PWL \(\bar{y}\), the metrics are
\begin{align}
\mathrm{MAE} &= \frac{1}{N}\sum_{i=1}^{N}\bigl|y_i-\widehat{y}_i\bigr|, \label{eq:mae}\\
\mathrm{RMSE} &= \sqrt{\frac{1}{N}\sum_{i=1}^{N}\bigl(y_i-\widehat{y}_i\bigr)^2}, \label{eq:rmse}\\
R^2 &= 1-\frac{\sum_{i=1}^{N}\bigl(y_i-\widehat{y}_i\bigr)^2}{\sum_{i=1}^{N}\bigl(y_i-\bar{y}\bigr)^2}, \label{eq:r2}\\
\mathrm{RTAE} &= 100\,\frac{\sum_{i=1}^{N}\bigl|y_i-\widehat{y}_i\bigr|}{\sum_{i=1}^{N}\bigl|y_i\bigr|}, \label{eq:rtae}\\
\delta_{\epsilon} &= \frac{100}{N}\sum_{i=1}^{N}\mathbb{1}\bigl[\bigl|y_i-\widehat{y}_i\bigr|>\epsilon\bigr], \qquad \epsilon\in\{0.1,0.5\}\,\mathrm{m}, \label{eq:metrics}\\
\mathrm{Acc}_0 &= 100\,\frac{\sum_{i=1}^{N}\mathbb{1}\bigl[y_i=0\wedge\widehat{y}_i=0\bigr]}{\sum_{i=1}^{N}\mathbb{1}\bigl[y_i=0\bigr]}. \label{eq:acc0}
\end{align}
Lower values are better for MAE, RMSE, RTAE, \(\delta_{0.1}\), and \(\delta_{0.5}\), and higher values are better for \(R^2\) and \(\mathrm{Acc}_0\). These metrics are used for reporting only, and the only quantity used for model selection is validation MSE.

\paragraph{Aggregation.}
Metrics are computed separately for each held-out scenario. For in-domain evaluation, scenario metrics are averaged within each seed, and we report the mean and sample standard deviation of the three seed-level means. For few-shot transfer, scenario metrics are computed separately for every support draw, averaged over all scenario and draw pairs within a seed so that each draw and scenario has equal weight, and then summarized per \(K\) as the mean and sample standard deviation over the three seeds. SLR transfer uses the same procedure. At \(K=0\), the single empty-support evaluation per seed replaces the repeated draws. No confidence intervals are reported, and no target test data are used for selection.

\begin{figure}[t]
    \centering
    \includegraphics[width=0.62\linewidth]{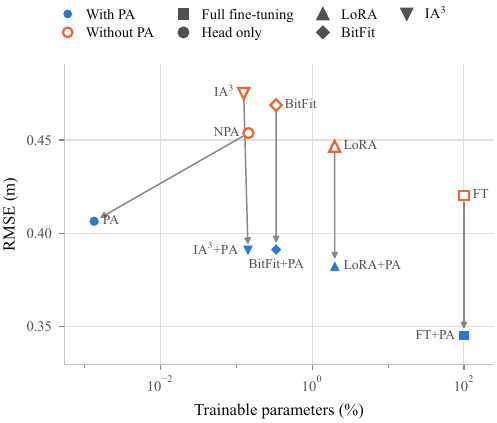}
    \caption{Mean transfer RMSE against the trainable parameter fraction of each adaptation regime. RMSE is averaged over the twelve backbones, the four transfer settings, and \(K\in\{0,1,3,5,10\}\), as in Table~\ref{tab:transfer_by_setting}, and each marker sits at the median trainable fraction over the backbones. Filled blue markers include the PA and open orange markers do not, while the marker shape gives the regime type. Each arrow joins a regime to its matched version with the PA. The horizontal axis is logarithmic.}
    \label{fig:complexity}
\end{figure}

\paragraph{Complexity profiling.}
For every backbone and regime, we record the total and trainable parameter counts, the frozen count as their difference, and the trainable fraction of Eq.~(\ref{eq:parameter_fraction}). These counts are exact and are read from the instantiated models. We do not compare FLOPs, MACs, or wall-clock times, since they could not be traced reliably for every architecture, and the runs used different GPUs (H100 or H200) and software environments. Figure~\ref{fig:complexity} plots the mean transfer RMSE of each regime against its trainable fraction. Adding the PA lowers the RMSE of every matched regime, by 17.9\% for full fine-tuning, 17.8\% for IA\(^3\), 16.5\% for BitFit, 14.3\% for LoRA, and 10.4\% for head-only adaptation. How many parameters this costs depends on the size of the backbone. In the dense-grid backbones the PA has 294 or 438 parameters, which is at most 0.08\% of the model and 50 to 160 times fewer than the raw head trained by NPA. The graph and operator backbones are much smaller, with about 18{,}000 to 131{,}000 parameters, so their PA of 4{,}358 parameters makes up 3.3\% to 23.8\% of the model. Across all backbones, adding the PA to a PEFT method raises the median trainable fraction by only 0.0033 percentage points, so the arrows in Figure~\ref{fig:complexity} are almost vertical. PA-only adaptation, which trains a median of 0.0012\% of the parameters, also reaches a lower mean RMSE than full fine-tuning without the PA (0.406\,m against 0.420\,m).

% ============================================================
% G. ADDITIONAL RESULTS  (add at the end of the appendix)
% ============================================================
\section{Additional Results}
\label{app:results}

\subsection{In-Domain Results}
\label{app:results_indomain}

Tables~\ref{tab:indomain_combined}--\ref{tab:indomain_sf} give all seven metrics of Appendix~\ref{app:evaluation_details} for in-domain prediction, averaged over both regions and for each region separately. Values are the mean and standard deviation over three seeds. For each backbone, the \emph{Source} column shows which source model (PA or raw) has the lower combined RMSE, and the same source model is reported in all three tables. Bold marks the best value in each column. VM-UNet is best on every metric in both regions. The dense backbones have lower MAE, RMSE, and error exceedance rates than the graph and operator backbones throughout, but ControlNet and ConvNeXt~V2 have a high and variable RTAE. This is likely because RTAE becomes unstable for scenarios with little flooding, where its denominator is small.

\begin{table*}[t]
\centering
\scriptsize
\setlength{\tabcolsep}{2pt}
\caption{In-domain results averaged over SF and AD.}
\label{tab:indomain_combined}

\resizebox{\textwidth}{!}{%
\begin{tabular}{l c c c c c c c c}
\hline
\textbf{Backbone}
& \textbf{Source}
& \textbf{MAE (m)}
& \textbf{RMSE (m)}
& \(\boldsymbol{R^2}\)
& \(\boldsymbol{\mathrm{Acc}_0}\) \textbf{(\%)}
& \textbf{RTAE (\%)}
& \(\boldsymbol{\delta_{0.5}}\) \textbf{(\%)}
& \(\boldsymbol{\delta_{0.1}}\) \textbf{(\%)} \\
\hline
VM-UNet & PA & \textbf{0.0026} $\pm$ 0.0001 & \textbf{0.0542} $\pm$ 0.0015 & \textbf{0.9660} $\pm$ 0.0008 & \textbf{99.9364} $\pm$ 0.0046 & \textbf{3.5310} $\pm$ 0.0806 & \textbf{0.0648} $\pm$ 0.0035 & \textbf{0.3044} $\pm$ 0.0230 \\
Swin~V2 & PA & 0.0037 $\pm$ 0.0002 & 0.0630 $\pm$ 0.0025 & 0.9582 $\pm$ 0.0055 & 99.8055 $\pm$ 0.0267 & 4.8003 $\pm$ 0.2261 & 0.1000 $\pm$ 0.0115 & 0.6650 $\pm$ 0.0454 \\
MaxViT & PA & 0.0038 $\pm$ 0.0003 & 0.0639 $\pm$ 0.0033 & 0.9564 $\pm$ 0.0039 & 99.7953 $\pm$ 0.0427 & 4.8321 $\pm$ 0.6176 & 0.0980 $\pm$ 0.0113 & 0.6697 $\pm$ 0.0680 \\
CASPIAN & PA & 0.0041 $\pm$ 0.0003 & 0.0686 $\pm$ 0.0036 & 0.9560 $\pm$ 0.0060 & 99.8178 $\pm$ 0.0077 & 5.5597 $\pm$ 0.4478 & 0.1266 $\pm$ 0.0161 & 0.7107 $\pm$ 0.0376 \\
Depth Pro & PA & 0.0046 $\pm$ 0.0007 & 0.0713 $\pm$ 0.0046 & 0.9522 $\pm$ 0.0005 & 99.7172 $\pm$ 0.0215 & 6.2101 $\pm$ 1.0633 & 0.1425 $\pm$ 0.0363 & 0.7708 $\pm$ 0.1173 \\
Depth Anything~V2 & PA & 0.0049 $\pm$ 0.0001 & 0.0747 $\pm$ 0.0017 & 0.9503 $\pm$ 0.0052 & 99.6402 $\pm$ 0.0806 & 6.7128 $\pm$ 0.1161 & 0.1561 $\pm$ 0.0092 & 0.8686 $\pm$ 0.0269 \\
ControlNet & PA & 0.0054 $\pm$ 0.0004 & 0.0754 $\pm$ 0.0030 & 0.9244 $\pm$ 0.0096 & 98.0455 $\pm$ 0.1461 & 15.4634 $\pm$ 7.0646 & 0.1618 $\pm$ 0.0124 & 1.0396 $\pm$ 0.1821 \\
ConvNeXt~V2 & PA & 0.0072 $\pm$ 0.0006 & 0.0893 $\pm$ 0.0028 & 0.9217 $\pm$ 0.0157 & 99.4124 $\pm$ 0.2542 & 26.5863 $\pm$ 13.7362 & 0.2307 $\pm$ 0.0139 & 1.4426 $\pm$ 0.3020 \\
\hline
MGN & PA & 0.0600 $\pm$ 0.0016 & 0.2990 $\pm$ 0.0034 & 0.9266 $\pm$ 0.0011 & 94.5578 $\pm$ 0.4711 & 9.0683 $\pm$ 0.3178 & 2.0276 $\pm$ 0.1528 & 8.3739 $\pm$ 0.2469 \\
GAT & Raw & 0.0617 $\pm$ 0.0021 & 0.3106 $\pm$ 0.0019 & 0.9235 $\pm$ 0.0007 & 96.7846 $\pm$ 0.6652 & 8.7774 $\pm$ 0.5934 & 1.9891 $\pm$ 0.0708 & 8.8130 $\pm$ 0.5266 \\
Transolver++ & Raw & 0.0649 $\pm$ 0.0078 & 0.3417 $\pm$ 0.0052 & 0.9215 $\pm$ 0.0079 & 96.5758 $\pm$ 4.0528 & 9.4857 $\pm$ 1.4286 & 1.7225 $\pm$ 0.3828 & 8.8934 $\pm$ 2.8365 \\
GCN & PA & 0.0737 $\pm$ 0.0025 & 0.3418 $\pm$ 0.0020 & 0.9121 $\pm$ 0.0009 & 93.5166 $\pm$ 0.7720 & 11.8118 $\pm$ 0.6456 & 1.6124 $\pm$ 0.0473 & 13.2854 $\pm$ 1.1726 \\
\hline
\end{tabular}%
}
\end{table*}

\begin{table*}[t]
\centering
\scriptsize
\setlength{\tabcolsep}{2pt}
\caption{In-domain results for Abu Dhabi.}
\label{tab:indomain_ad}

\resizebox{\textwidth}{!}{%
\begin{tabular}{l c c c c c c c c}
\hline
\textbf{Backbone}
& \textbf{Source}
& \textbf{MAE (m)}
& \textbf{RMSE (m)}
& \(\boldsymbol{R^2}\)
& \(\boldsymbol{\mathrm{Acc}_0}\) \textbf{(\%)}
& \textbf{RTAE (\%)}
& \(\boldsymbol{\delta_{0.5}}\) \textbf{(\%)}
& \(\boldsymbol{\delta_{0.1}}\) \textbf{(\%)} \\
\hline
VM-UNet & PA & \textbf{0.0039} $\pm$ 0.0002 & \textbf{0.0866} $\pm$ 0.0030 & \textbf{0.9469} $\pm$ 0.0015 & \textbf{99.9012} $\pm$ 0.0098 & \textbf{4.4593} $\pm$ 0.1712 & \textbf{0.1058} $\pm$ 0.0095 & \textbf{0.4604} $\pm$ 0.0384 \\
Swin~V2 & PA & 0.0044 $\pm$ 0.0002 & 0.0902 $\pm$ 0.0014 & 0.9450 $\pm$ 0.0013 & 99.7979 $\pm$ 0.0659 & 5.5141 $\pm$ 0.0702 & 0.1175 $\pm$ 0.0084 & 0.7135 $\pm$ 0.0074 \\
MaxViT & PA & 0.0048 $\pm$ 0.0004 & 0.0943 $\pm$ 0.0045 & 0.9431 $\pm$ 0.0034 & 99.7823 $\pm$ 0.0865 & 5.8221 $\pm$ 0.6174 & 0.1261 $\pm$ 0.0105 & 0.7930 $\pm$ 0.0977 \\
CASPIAN & PA & 0.0055 $\pm$ 0.0007 & 0.0984 $\pm$ 0.0061 & 0.9412 $\pm$ 0.0034 & 99.7413 $\pm$ 0.0271 & 7.0318 $\pm$ 0.9230 & 0.1648 $\pm$ 0.0330 & 0.9159 $\pm$ 0.1249 \\
Depth Pro & PA & 0.0064 $\pm$ 0.0015 & 0.1099 $\pm$ 0.0106 & 0.9328 $\pm$ 0.0103 & 99.7026 $\pm$ 0.0619 & 8.4506 $\pm$ 2.3253 & 0.2215 $\pm$ 0.0785 & 0.9217 $\pm$ 0.2590 \\
Depth Anything~V2 & PA & 0.0069 $\pm$ 0.0005 & 0.1143 $\pm$ 0.0055 & 0.9293 $\pm$ 0.0036 & 99.5482 $\pm$ 0.1792 & 9.3712 $\pm$ 0.1613 & 0.2330 $\pm$ 0.0247 & 1.0656 $\pm$ 0.0793 \\
ControlNet & PA & 0.0061 $\pm$ 0.0007 & 0.1025 $\pm$ 0.0059 & 0.9148 $\pm$ 0.0150 & 98.0762 $\pm$ 0.2425 & 14.4761 $\pm$ 11.2414 & 0.1907 $\pm$ 0.0175 & 1.0372 $\pm$ 0.1651 \\
ConvNeXt~V2 & PA & 0.0076 $\pm$ 0.0010 & 0.1140 $\pm$ 0.0074 & 0.9155 $\pm$ 0.0271 & 99.4878 $\pm$ 0.4062 & 23.5902 $\pm$ 22.2227 & 0.2613 $\pm$ 0.0251 & 1.3143 $\pm$ 0.2378 \\
\hline
MGN & PA & 0.0664 $\pm$ 0.0023 & 0.3734 $\pm$ 0.0068 & 0.9134 $\pm$ 0.0019 & 95.0695 $\pm$ 0.3636 & 10.2942 $\pm$ 0.3939 & 2.2620 $\pm$ 0.0900 & 8.1504 $\pm$ 0.3404 \\
GAT & Raw & 0.0704 $\pm$ 0.0021 & 0.3881 $\pm$ 0.0029 & 0.9106 $\pm$ 0.0013 & 96.4951 $\pm$ 0.8523 & 10.3491 $\pm$ 0.8927 & 2.5371 $\pm$ 0.0782 & 8.5326 $\pm$ 0.7478 \\
Transolver++ & Raw & 0.0834 $\pm$ 0.0163 & 0.4557 $\pm$ 0.0072 & 0.8960 $\pm$ 0.0141 & 93.9065 $\pm$ 8.0915 & 12.4061 $\pm$ 2.5507 & 2.4486 $\pm$ 0.7725 & 9.9514 $\pm$ 5.7027 \\
GCN & PA & 0.0802 $\pm$ 0.0008 & 0.4283 $\pm$ 0.0027 & 0.8959 $\pm$ 0.0011 & 94.1315 $\pm$ 0.5815 & 12.9014 $\pm$ 0.3223 & 2.0481 $\pm$ 0.0692 & 11.5205 $\pm$ 0.4503 \\
\hline
\end{tabular}%
}
\end{table*}

\begin{table*}[t]
\centering
\scriptsize
\setlength{\tabcolsep}{2pt}
\caption{In-domain results for San Francisco.}
\label{tab:indomain_sf}

\resizebox{\textwidth}{!}{%
\begin{tabular}{l c c c c c c c c}
\hline
\textbf{Backbone} 
& \textbf{Source} 
& \textbf{MAE (m)} 
& \textbf{RMSE (m)} 
& \(\boldsymbol{R^2}\) 
& \(\boldsymbol{\mathrm{Acc}_0}\) \textbf{(\%)} 
& \textbf{RTAE (\%)} 
& \(\boldsymbol{\delta_{0.5}}\) \textbf{(\%)} 
& \(\boldsymbol{\delta_{0.1}}\) \textbf{(\%)} \\
\hline
VM-UNet & PA & \textbf{0.0013} $\pm$ 0.0000 & \textbf{0.0217} $\pm$ 0.0008 & \textbf{0.9851} $\pm$ 0.0002 & \textbf{99.9715} $\pm$ 0.0028 & \textbf{2.6026} $\pm$ 0.0221 & \textbf{0.0239} $\pm$ 0.0024 & \textbf{0.1484} $\pm$ 0.0133 \\
Swin~V2 & PA & 0.0030 $\pm$ 0.0005 & 0.0357 $\pm$ 0.0050 & 0.9714 $\pm$ 0.0106 & 99.8130 $\pm$ 0.0193 & 4.0864 $\pm$ 0.3839 & 0.0824 $\pm$ 0.0268 & 0.6164 $\pm$ 0.0879 \\
MaxViT & PA & 0.0027 $\pm$ 0.0002 & 0.0335 $\pm$ 0.0021 & 0.9697 $\pm$ 0.0044 & 99.8083 $\pm$ 0.0133 & 3.8421 $\pm$ 0.6210 & 0.0698 $\pm$ 0.0126 & 0.5464 $\pm$ 0.0425 \\
CASPIAN & PA & 0.0028 $\pm$ 0.0001 & 0.0387 $\pm$ 0.0012 & 0.9708 $\pm$ 0.0091 & 99.8943 $\pm$ 0.0297 & 4.0875 $\pm$ 0.0611 & 0.0885 $\pm$ 0.0010 & 0.5055 $\pm$ 0.0569 \\
Depth Pro & PA & 0.0028 $\pm$ 0.0002 & 0.0327 $\pm$ 0.0021 & 0.9716 $\pm$ 0.0101 & 99.7319 $\pm$ 0.0192 & 3.9696 $\pm$ 0.2843 & 0.0636 $\pm$ 0.0088 & 0.6199 $\pm$ 0.0340 \\
Depth Anything~V2 & PA & 0.0030 $\pm$ 0.0003 & 0.0352 $\pm$ 0.0022 & 0.9712 $\pm$ 0.0097 & 99.7323 $\pm$ 0.0281 & 4.0544 $\pm$ 0.2756 & 0.0793 $\pm$ 0.0114 & 0.6716 $\pm$ 0.1031 \\
ControlNet & PA & 0.0047 $\pm$ 0.0006 & 0.0483 $\pm$ 0.0025 & 0.9340 $\pm$ 0.0079 & 98.0148 $\pm$ 0.0794 & 16.4508 $\pm$ 5.3722 & 0.1328 $\pm$ 0.0077 & 1.0420 $\pm$ 0.2952 \\
ConvNeXt~V2 & PA & 0.0069 $\pm$ 0.0010 & 0.0646 $\pm$ 0.0029 & 0.9279 $\pm$ 0.0116 & 99.3370 $\pm$ 0.1479 & 29.5824 $\pm$ 10.2942 & 0.2001 $\pm$ 0.0029 & 1.5708 $\pm$ 0.5574 \\
\hline
MGN & PA & 0.0535 $\pm$ 0.0010 & 0.2247 $\pm$ 0.0015 & 0.9398 $\pm$ 0.0008 & 94.0462 $\pm$ 0.5802 & 7.8424 $\pm$ 0.4243 & 1.7932 $\pm$ 0.2360 & 8.5974 $\pm$ 0.1784 \\
GAT & Raw & 0.0530 $\pm$ 0.0022 & 0.2330 $\pm$ 0.0009 & 0.9364 $\pm$ 0.0009 & 97.0741 $\pm$ 0.6491 & 7.2056 $\pm$ 0.6281 & 1.4410 $\pm$ 0.1047 & 9.0934 $\pm$ 0.3633 \\
Transolver++ & Raw & 0.0463 $\pm$ 0.0022 & 0.2277 $\pm$ 0.0036 & 0.9469 $\pm$ 0.0030 & 99.2451 $\pm$ 0.0141 & 6.5653 $\pm$ 0.3107 & 0.9964 $\pm$ 0.0152 & 7.8354 $\pm$ 0.4394 \\
GCN & PA & 0.0671 $\pm$ 0.0042 & 0.2552 $\pm$ 0.0012 & 0.9283 $\pm$ 0.0007 & 92.9016 $\pm$ 0.9625 & 10.7222 $\pm$ 0.9688 & 1.1767 $\pm$ 0.0254 & 15.0503 $\pm$ 1.8948 \\
\hline
\end{tabular}%
}
\end{table*}

\subsection{Per-Backbone Transfer Results}
\label{app:results_transfer}

\begin{figure}[t]
    \centering
    \includegraphics[width=\linewidth]{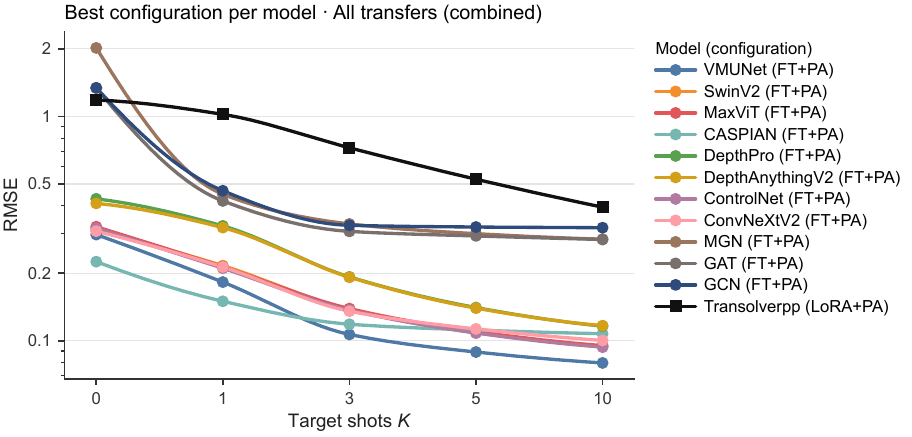}
    \caption{Best regime per backbone over all transfer settings, selected by mean RMSE over \(K\). The vertical axis is logarithmic.}
    \label{fig:best_regime_all}
\end{figure}

\begin{figure}[t]
    \centering
    \includegraphics[width=\linewidth]{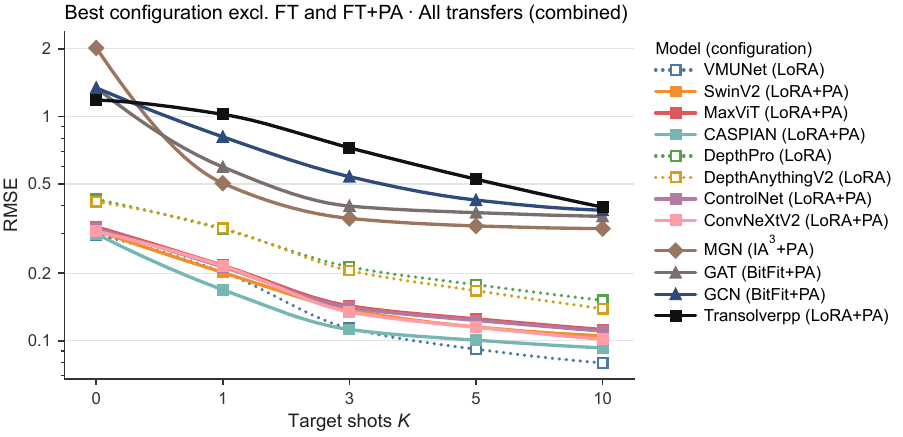}
    \caption{Best regime per backbone when FT and FT+PA are excluded. Solid lines are regimes with the PA and dotted lines are regimes without it. The vertical axis is logarithmic.}
    \label{fig:best_regime_noft}
\end{figure}

Figures~\ref{fig:best_regime_all} and~\ref{fig:best_regime_noft} show, for each backbone, the RMSE curve of its best regime over all four transfer settings, first over all ten regimes and then with the two full fine-tuning regimes excluded. The best regime is the one with the lowest RMSE averaged over \(K\in\{0,1,3,5,10\}\), with equal weight for each \(K\). In Figure~\ref{fig:best_regime_all}, FT+PA is best for every backbone except Transolver++, where LoRA+PA is best. The dense and depth-foundation backbones reach RMSE close to 0.1\,m by \(K=10\), while the graph and operator backbones start from a much higher zero-shot error and level off near 0.3\,m. When full fine-tuning is excluded (Figure~\ref{fig:best_regime_noft}), nine of the twelve backbones still select a PA regime. Table~\ref{tab:transfer_by_setting} breaks the regime comparison down by transfer setting.

\begin{table}[t]
\centering
\small
\setlength{\tabcolsep}{4pt}
\caption{RMSE (m) of each regime averaged over all twelve backbones and over \(K\in\{0,1,3,5,10\}\), for each transfer setting. PEFT rows give the mean of LoRA, IA\(^3\) and BitFit, with the individual methods listed beneath. Bold marks the best regime in each column.}
\label{tab:transfer_by_setting}
\begin{tabular}{l c c c c}
\hline
\textbf{Regime} & \textbf{SF\(\rightarrow\)AD} & \textbf{AD\(\rightarrow\)SF} & \textbf{SF\(_{1.0}\rightarrow\)SF\(_{0.5}\)} & \textbf{SF\(_{1.0}\rightarrow\)SF\(_{1.5}\)} \\
\hline
FT+PA & \textbf{0.4338} & \textbf{0.5401} & \textbf{0.1668} & \textbf{0.2403} \\
\emph{PEFT+PA (mean)} & \emph{0.4989} & \emph{0.5992} & \emph{0.1819} & \emph{0.2727} \\
\quad LoRA+PA & 0.5233 & 0.5653 & 0.1793 & 0.2629 \\
\quad IA\(^3\)+PA & 0.4847 & 0.6170 & 0.1818 & 0.2785 \\
\quad BitFit+PA & 0.4887 & 0.6152 & 0.1847 & 0.2766 \\
PA & 0.5093 & 0.6198 & 0.1969 & 0.2998 \\
\hline
FT & 0.6106 & 0.6233 & 0.1786 & 0.2686 \\
\emph{PEFT (mean)} & \emph{0.6800} & \emph{0.7061} & \emph{0.1859} & \emph{0.2817} \\
\quad LoRA & 0.6254 & 0.7122 & 0.1781 & 0.2708 \\
\quad IA\(^3\) & 0.6891 & 0.7286 & 0.1913 & 0.2905 \\
\quad BitFit & 0.7256 & 0.6774 & 0.1882 & 0.2836 \\
NPA & 0.6902 & 0.6243 & 0.1973 & 0.3030 \\
\hline
\end{tabular}
\end{table}

\subsection{Qualitative Error Maps}
\label{app:results_qualitative}

\begin{figure*}[t]
    \centering
    \includegraphics[width=\linewidth]{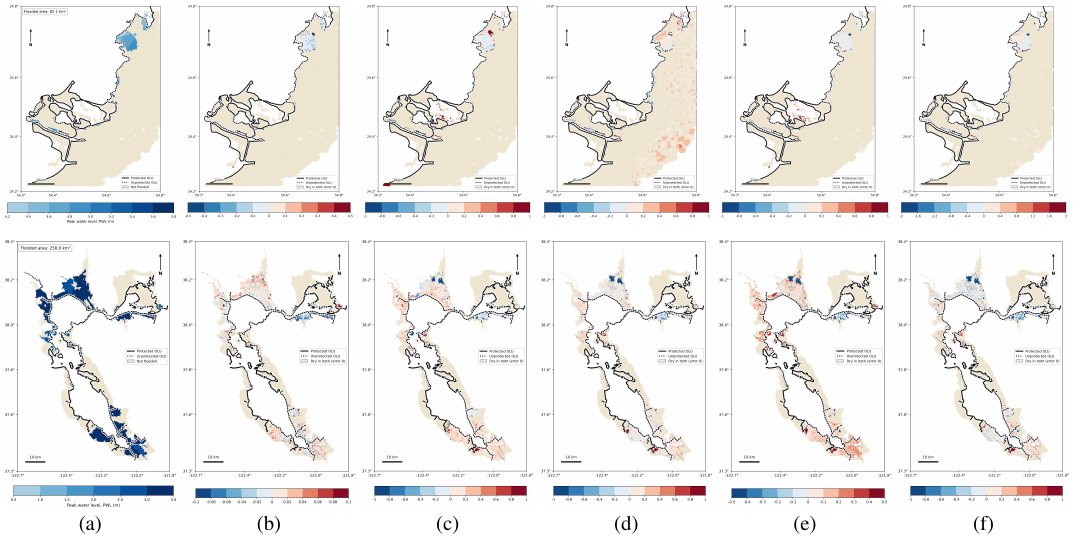}
    \caption{PWL error maps with VM-UNet (prediction minus ground truth, in m) for the scenario and regimes of Figure~\ref{fig:qualitative}, AD in the top row and SF in the bottom row. (a)~Ground-truth PWL for reference. (b)~In-domain error. (c)--(f)~Errors after transfer with \(K=3\) using (c)~FT+PA, (d)~FT, (e)~best PEFT+PA, and (f)~best PEFT. Beige marks locations that are dry in both the prediction and the ground truth. Color scales differ between panels.}
    \label{fig:qualitative_error}
\end{figure*}

Figure~\ref{fig:qualitative_error} shows the VM-UNet error maps for the scenario and regimes of Figure~\ref{fig:qualitative}, computed as prediction minus ground truth at each location, so red marks overestimated PWL and blue marks underestimated PWL. Locations that are dry in both the prediction and the ground truth are shown in beige. In AD, the errors of FT and LoRA are spread over a large inland area, which matches the false flooding seen in Figure~\ref{fig:qualitative}. The PA regimes keep the errors close to the coast, where the true flooding occurs. In SF, the in-domain errors stay within \(\pm0.1\)\,m. After transfer, the largest errors appear in the northern basin, where FT and LoRA underestimate PWL, and in a few small areas along the southern shoreline.

\end{document}